\documentclass[11pt]{article}

\usepackage{setspace,graphicx,epstopdf,amsmath,amsfonts,amssymb,amsthm,nccmath,versionPO,mathrsfs}
\usepackage{marginnote,datetime,enumitem,rotating,fancyvrb}
\usepackage{float}
\usepackage{titling}
\usepackage{xpatch}
\usepackage{subcaption}
\usepackage{authblk} 

\makeatletter
{\begin{titlepage}\def\@thanks{}}%
	{\end{titlepage}}
\xpatchcmd\titlepage{\setcounter{page}\@ne}{}{}{}
\xpatchcmd\endtitlepage{\setcounter{page}\@ne}{}{}{}
\makeatother

\usepackage[stable,bottom]{footmisc}
\usepackage[dvipsnames]{xcolor}
\usepackage{hyperref}
\usepackage[capitalize]{cleveref}
\hypersetup{
	colorlinks  = true,
	urlcolor    = blue,
	linkcolor   = red,
	citecolor   = blue
}

\usepackage{microtype}

\usepackage[authoryear,round]{natbib}
\usdate

\usepackage{titlesec}
\newcommand{\periodafter}[1]{#1.}
\titleformat{\subsubsection}[runin]
{\normalfont\bfseries}{\thesubsubsection}{1em}{\periodafter}

\excludeversion{notes}		
\includeversion{links}      

\iflinks{}{\hypersetup{draft=true}}

\ifnotes{%
	\usepackage[margin=1in,paperwidth=10in,right=2.5in]{geometry}%
	\usepackage[textwidth=1.4in,shadow,colorinlistoftodos]{todonotes}%
}{%
	\usepackage[margin=1in]{geometry}%
	\usepackage[disable]{todonotes}%
}

\makeatletter\let\chapter\@undefined\makeatother 

\theoremstyle{definition}

\usepackage{booktabs}
\usepackage{bbm}
\usepackage{dsfont}
\usepackage{multirow}
\usepackage{tabularx,booktabs,caption}
\usepackage[flushleft]{threeparttable}
\usepackage[title]{appendix}
\usepackage{lscape}
\usepackage{longtable,rotating}
\usepackage{rotating}
\usepackage{bm,bbm}
\usepackage{mathtools}
\usepackage[font={onehalfspacing},labelfont=bf]{caption}
\usepackage[title]{appendix}
\usepackage{siunitx}
\usepackage{scrextend}
\usepackage{longtable}
\usepackage{slashbox}
\usepackage{rotating}
\mathtoolsset{centercolon}
\usepackage{tikz}
\usetikzlibrary{positioning,arrows.meta,shapes.geometric}
\usepackage{pdflscape}
\usepackage{adjustbox}
\everymath{\displaystyle}

\newcolumntype{Y}{>{\raggedleft\arraybackslash}X}

\usepackage{setspace}
\usepackage{authblk}
\usepackage{hyperref}
\usepackage{soul}

\usepackage{graphicx}
\usepackage{subcaption}
\makeatletter
\newcounter{mysubfig}
\renewcommand{\themysubfig}{\alph{mysubfig}}
\newcommand{\mysubcaption}[1]{%
  \refstepcounter{mysubfig}%
  \par\footnotesize\centering(\themysubfig)\space #1\par
}

\usepackage{ifthen}

\newcounter{noteGlobalCtr}
\definecolor{colour3}{RGB}{178,55,250} 
\definecolor{colour1}{RGB}{255,130,20} 

\definecolor{darkpastelgreen}{rgb}{0.01, 0.75, 0.24}

\newcommand{\MC}[1]{}
\newcommand{\MG}[1]{}

\newcommand{\mc}[1]{#1}
\newcommand{\mcc}[1]{#1}

\newcommand{\mg}[1]{#1}

\makeatother

\begin{document}
\doublespacing

\title{\textbf{A Bipartite Graph Approach to U.S.-China Cross-Market Return Forecasting}}

\author[1]{Jing Liu\thanks{Corresponding author; Email: jing.liu@exeter.ox.ac.uk}}
\author[2]{Maria Grith}
\author[3]{Xiaowen Dong}
\author[4,1,5]{Mihai Cucuringu}

\affil[1]{Department of Statistics, University of Oxford, UK}
\affil[2]{Finance Department, Neoma Business School, France}
\affil[3]{Department of Engineering Science, University of Oxford, UK}
\affil[4]{Department of Mathematics, University of California Los Angeles, US}
\affil[5]{Oxford-Man Institute of Quantitative Finance, University of Oxford, UK}

\date{}

\maketitle

\begin{abstract}
This paper studies cross-market return predictability through a machine learning framework that preserves economic structure. Exploiting the non-overlapping trading hours of the U.S. and Chinese equity markets, we construct a directed bipartite graph that captures time-ordered predictive linkages between stocks across markets. Edges are selected via rolling-window hypothesis testing, and the resulting graph serves as a sparse, economically interpretable feature-selection layer for downstream machine learning models. We apply a range of regularized and ensemble methods to forecast open-to-close returns using lagged foreign-market information. Our results reveal a pronounced directional asymmetry: U.S. previous-close-to-close returns contain substantial predictive information for Chinese intraday returns, whereas the reverse effect is limited. This informational asymmetry translates into economically meaningful performance differences and highlights how structured machine learning frameworks can uncover cross-market dependencies while maintaining interpretability.
\end{abstract}


\noindent \mg{\textbf{JEL Classification:} G17, G15, C58}

\vspace{1cm}



\MG{Alternative title: A Bipartite Graph Approach to U.S.–China Stock Forecasting}
\MC{Indeed, would be good to have Bipartite in the title, makes it a bit more. I like the above; other options would be } 

\MC{Directed Bipartite Graph Feature Selection for U.S.--China Cross-Market Return Forecasting with Machine Learning}

\MC{Directed Bipartite Graphs for U.S.--China Cross-Market Return Forecasting}

\section{Introduction}\label{sec1}

\mcc{Return prediction remains a central problem in empirical asset pricing and portfolio management, yet its statistical difficulty is amplified by noise, non-stationarity, and nonlinear dependence structures in financial markets. While machine learning methods have become increasingly prevalent in single-market forecasting applications \citep{chen2015lstm, wang2024stock, yang2026enhancing}, comparatively little attention has been paid to stock-level cross-market return prediction under realistic trading-session timing constraints.}

Most existing studies on return forecasting focus on predicting within a single market. For example, \citet{chen2015lstm} apply a Long Short-Term Memory (LSTM) model to predict stock returns in the Chinese market, while \citet{wang2024stock} studies U.S. stock return prediction using neural network models. Similarly, \citet{yang2026enhancing} propose an intraday volume-based uncertainty proxy to predict return direction in the Chinese market. \mcc{These studies demonstrate the growing use of machine learning methods in single-market settings.}
%
By contrast, research on cross-market interactions has largely emphasized contemporaneous co-movement, spillovers, or causal transmission rather than explicit stock-level return prediction. 
For instance, \citet{eun1989international} analyze the international transmission of stock market movements using vector autoregression, \citet{baur2006return} evaluate contemporaneous return correlations using GARCH models, and \citet{rapach2013international} document the leading role of the U.S. market through causality tests. \citet{sarwar2014us} examine the relationship between U.S. market uncertainty and European equity returns during crisis periods, while \citet{jung2024threshold} study interdependency patterns between the U.S. and Chinese markets using threshold overnight co-movement processes.

\mcc{Only a limited number of studies have attempted explicit cross-market predictive analysis using machine learning models. For example, \citet{lee2020multimodal} apply a deep neural network to fuse information from the U.S. and South Korean markets for index-level return prediction, and \citet{kumar2024dynamic} propose a graph neural network to model volatility spillovers across markets. However, most existing work operates at the index level, and to our knowledge no prior study has examined stock-level cross-market return prediction between the U.S. and Chinese markets under realistic trading-session timing.}


\mcc{Our study fills this gap by developing a directed bipartite graph framework for stock-level cross-market return forecasting between the U.S. and Chinese equity markets. We construct a time-ordered bipartite graph that selects cross-market predictors based on rolling-window screening, thereby capturing directed predictive links across non-overlapping trading sessions. The selected predictors are then embedded into a suite of ten machine learning models to forecast next-session open-to-close (OPCL) returns in each market.} 

\mcc{Empirically, we demonstrate a pronounced directional asymmetry: U.S. market information is substantially more informative for predicting Chinese stock returns than vice versa. Sharpe Ratios (SRs) obtained when forecasting Chinese stocks using U.S. predictors consistently exceed those in the reverse direction. We further show that both the graph-based selection mechanism and cross-market information contribute materially to predictive performance.}

\mcc{Sector-level patterns in the estimated graph reveal economically interpretable transmission channels across markets. \mg{For instance, sector-level aggregation of the bipartite graph reveals meaningful cross-sector transmission patterns rather than a block-diagonal structure.} Our focus is on documenting directional cross-market predictability and the structure of the associated dependency graph rather than designing a fully implementable trading strategy. Accordingly, performance metrics are reported pre-transaction-cost and without liquidity-optimized weighting, and should be interpreted as evidence of predictive asymmetry rather than deployable alpha.}

\mcc{Our setting is economically and statistically distinctive because the U.S. and Chinese equity sessions do not overlap. This implies that U.S. previous-close-to-close (pvCLCL) information is fully observed before the subsequent Chinese OPCL window begins, yielding a clean timing structure for cross-market prediction. The directed bipartite graph can therefore be interpreted as a time-ordered map of potential information transmission channels across markets, rather than a contemporaneous correlation network.}

The structure of the paper is arranged as follows. Section~\ref{sec:literature} provides a detailed review of related work. Section~\ref{sec:data} describes the data we use and the definitions of financial terms involved. Section~\ref{sec:method} introduces the graph-based methodology for feature selection and prediction. Section~\ref{sec:exp} presents the evaluation metrics and experimental results. 
Finally, Section~\ref{sec:end} summarizes the study and discusses future research directions.

\section{Related Literature}
\label{sec:literature}

\MC{Think about where to move the Kumar citation from the intro, as it didn't sort of belong there; it was this sentence:} 

\subsection{Cross-Market Analysis and Prediction}
\mcc{Global financial markets have become increasingly interconnected with the intensification of international economic and financial integration. As a result, shocks, volatility, and information can propagate rapidly across countries through multiple transmission channels. A substantial body of research therefore examines cross-market linkages, including price discovery, return co-movement, volatility spillovers, and broader measures of financial interconnectedness, typically within econometric frameworks. 
Such interconnectedness has motivated studies of directional information flow and market leadership across countries.}
For example, \citet{liu2011information} examine information transmission and price discovery between the U.S. and Chinese markets. \citet{asgharian2013spatial} study how economic and geographical relationships across countries affect stock market returns. \citet{mohammadi2015return} analyze daily returns and volatility dynamics in the U.S. and Chinese markets. \citet{clements2015volatility} investigate global transmission of news and volatility across financial markets, while \citet{ahmad2018financial} explore market interconnectedness through return and volatility spillovers. \citet{huang2023cross} construct a financial network to characterize cross-market risk spillovers and interaction topology. These studies primarily emphasize contemporaneous relationships and transmission mechanisms rather than explicit stock-level return prediction.

Beyond studying cross-market information propagation, \mcc{a growing strand of research incorporates signals} from multiple markets into forecasting models to improve predictive performance. Such integration typically relies on feature engineering that embeds external market indicators, deep learning architectures that fuse multi-market inputs, or graph-based models designed to capture inter-market dependencies. For example, \citet{thenmozhi2016forecasting} use foreign index information to enhance index prediction, \citet{lee2020multimodal} develop multimodal deep learning models for cross-market index forecasting, and \citet{lin2025cspo} leverage external futures market data to predict movements of the China Securities Index. \citet{gong2025cross} propose a cross-market volatility forecasting framework exploiting risk transmission across markets. 

\mcc{However, much of this literature focuses on aggregate indices or volatility measures rather than stock-level return prediction. Moreover, network structures are often employed to characterize spillovers and interconnectedness rather than as predictive screening devices for individual stocks. Explicit stock-level cross-market return forecasting under time-ordered, non-overlapping trading sessions remains largely unexplored.}
\mcc{Applying our methodology in this setting is therefore novel. Empirically, the directed bipartite structure reveals pronounced asymmetry in cross-market predictability, with U.S. stocks exerting substantially stronger predictive influence on Chinese stocks than vice versa.}
\mg{These findings align with the literature documenting asymmetric cross-market return predictability with a leading role for the U.S.. \citet{rapach2013international} show that lagged U.S. market returns possess substantial predictive power for foreign equity markets, while the reverse predictability is considerably weaker, highlighting the central role of the U.S. in global price discovery. Similarly, \citet{siliverstovs2017international} finds that the predictive influence of the U.S. is particularly pronounced during market downturns, reinforcing the view that U.S. information dominates international return dynamics. Focusing specifically on China-related markets, \citet{mohammadi2015return} document significant return and volatility spillovers from the U.S. to Chinese mainland and Hong Kong, with weaker effects in the opposite direction.}

\MG{@Jing: I like that you cited several papers focusing on China - US interactions. Could you give a brief idea about what these papers find w.r.t. the direction of the interactions? Then we could say that this studies supports or not earlier studies. How does this paper contribute to the empirical evidence? DONE}

\subsection{Graph Methods in Finance}
Graph methods provide a way to represent relationships among financial entities, rather than treating each entity in isolation. The use of graphs aligns with the view that financial systems are interconnected~\citep{bardoscia2021physics}, and that modeling these interconnections can improve forecasting and risk‐management~\citep{chen2025financial}. Many financial phenomena, such as asset co-movements, spillovers and supply-chain linkages, are naturally represented as graphs.

\mcc{Bipartite graphs, which originate in graph theory and network science as representations of relationships between two distinct sets of nodes \citep{guillaume2006bipartite, newman2018networks}, provide a natural framework for modeling interactions across disjoint groups. In economics and finance, bipartite structures arise in contexts such as credit networks, production networks, and supply-chain relationships, where connections form between two heterogeneous sets of entities rather than within a single homogeneous market.} \mg{For instance, \cite{kley2020modelling} study extremal dependence for operational risk by a bipartite graph. \citet{wang2020bipartite} design a bipartite-graph-based recommender for crowdfunding with sparse data. In econometrics, \cite{wu2024bipartite} propose a quasi-maximum likelihood approach to estimate a bipartite network influence model.}

\mcc{A growing literature applies graph neural networks (GNNs) and related architectures to financial forecasting tasks.} 
\citet{wang2021review} provide a survey of GNN methods in financial applications, including stock movement prediction, loan default risk assessment, recommender systems, fraud detection, and other financial events. \citet{chen2018incorporating} apply a Graph Convolutional Network (GCN) to integrate information from related companies and improve stock price prediction. \citet{li2021modeling} propose an LSTM-relational GCN that captures inter-stock relationships through correlation matrices to predict overnight movements. \citet{capponi2024graph} develop a GNN framework for asset pricing using supply-chain data. \citet{zhang2025forecasting} incorporate cross-stock spillover effects to forecast multivariate realized volatilities, and \citet{luo2025spatial} construct a semantic company relationship graph to enhance stock price forecasting.

Some recent empirical work extends graph-based forecasting to richer and more dynamic architectures. For example, \citet{cheng2021modeling} employ a Graph Attention Network (GAT) to model momentum spillovers in stock returns, while \citet{kumar2024dynamic} introduce a temporal GAT that combines graph convolution and attention mechanisms to capture structural and temporal dependencies across global market indices. \citet{lee2025symmetry} show that GCN- and GAT-based models can outperform conventional machine learning baselines by exploiting symmetric interdependencies among financial indices. Related research also incorporates multimodal information into graph construction. \citet{cheng2022financial} integrate financial events and news into a multimodal GNN framework for price prediction, and \citet{liu2024multimodal} develop a multiscale dynamic GCN that combines textual and numerical inputs to forecast stock movements.

\mcc{Despite these advances, most graph-based forecasting models construct within-market networks, where edges are defined through contemporaneous similarity, correlation, or learned attention mechanisms. Such graphs typically capture symmetric interdependencies among assets within a single market and are primarily used to enhance predictive performance through richer representation learning. In contrast, our framework constructs a directed bipartite graph across two distinct markets, where edges are formed through time-ordered predictive screening rather than contemporaneous association. The resulting graph serves as a feature-selection mechanism for stock-level cross-market return prediction, explicitly exploiting the non-overlapping trading sessions between the U.S. and Chinese markets.} 


\MG{@Jing: \\
(1) What are the primary references for bipartite graphs and in which areas they emerged? (keep it short, 1-2 sentences. DONE) \MC{just added standard Newman 2010 textbook reference for networks. And also this \cite{GUILLAUME_Bipartite} from  \url{https://www.sciencedirect.com/science/article/abs/pii/S0378437106004638} -- seems highly cited, but from the physics literature.} \\
(2) Add a few references on bipartite graphs in econ, finance. Try to explain to which extent what we do is new or different than the earlier papers. (I sent you a list of relevant journals in finance/econometrics) Write one sentence to what we find and how this contributes to the literature. DONE}

\MC{@Maria, if you paste the  above list here, I can also help and have a look at some of the references.} \MG{I added the list of econ/finance journals and commented it out below. DONE.}


\subsection{Machine Learning in Finance}
Driven by increasing data availability and computational power, the application of machine learning \mcc{in finance has expanded substantially} in recent years. Compared to classical time-series and econometric models, such as ARIMA and GARCH, machine learning approaches \mcc{are often considered better suited to high-dimensional and nonlinear settings}. A survey by~\citet{rundo2019machine} \mcc{documents that} machine-learning-based systems demonstrate superior overall performance compared to traditional approaches. Another survey by~\citet{kelly2023financial} \mcc{highlights how machine learning methods have become established in empirical financial research}. Key applications include forecasting asset returns, volatility estimation, fraud detection, and algorithmic trading. \mcc{Forecasting asset returns remains inherently difficult due to low signal-to-noise ratios, structural instability, and nonlinear dependence patterns. Moreover, evidence of predictive gains is often sensitive to model specification and feature construction. These challenges partly motivate the adoption of flexible machine learning methods and the incorporation of richer information sets.} Given recent developments in machine learning, its applications in finance can be grouped into several major categories: traditional machine learning methods, deep learning methods, and large-language-model-based methods.

For traditional machine learning methods,~\citet{huang2005forecasting} \mcc{employ} support vector machines (SVM) to predict the direction of weekly price movements.~\citet{kumar2006forecasting} \mcc{investigate} the application of SVM and Random Forests (RF) in predicting the direction of a market index.~\citet{cakra2015stock} \mcc{incorporate} sentiment information and \mcc{use} a basic linear regression model for stock price prediction.~\citet{thenmozhi2016forecasting} \mcc{predict} stock prices of several major indices using support vector regression.~\citet{yang2026enhancing} \mcc{propose} a novel proxy and \mcc{apply} Extreme Gradient Boosting (XGBoost) to predict return directions in the Chinese market.

For deep learning methods,~\citet{chen2015lstm} \mcc{use} an LSTM model for sequence learning and Chinese stock return forecasting.~\citet{wang2024stock} \mcc{investigates} the performance of neural network models in predicting stock returns. A survey by~\citet{gao2024machine} \mcc{highlights} the expanding use of deep neural networks, convolutional neural networks, recurrent neural networks, and other advanced architectures in financial contexts.

For large-language-model-based methods,~\citet{nie2024survey} \mcc{review} how Large Language Models (LLMs) \mcc{are applied} in finance.~\citet{ding2023integrating} \mcc{demonstrate} the effectiveness of LLMs in forecasting stock returns.~\citet{chen2023chatgpt} \mcc{propose} a framework that \mcc{integrates} ChatGPT and GNN to forecast stock movements.~\citet{chen2024does} \mcc{investigate} the ability of ChatGPT for stock return forecasting.

\mcc{Despite these advances, most existing studies focus on single-market return prediction and rely primarily on information drawn from within the same market. Additional information is often shown to improve predictive performance, yet it is typically incorporated in contemporaneous or symmetric settings. Very few studies employ machine learning methods in stock-level cross-market forecasting environments characterized by asynchronous trading sessions and explicitly time-ordered information flows. Differing from existing studies, our framework combines directed bipartite screening with a second-stage machine learning prediction step, enabling systematic exploitation of cross-market dependencies, temporal ordering, and asymmetric predictive structure.}

\MG{@Jing: I understand that this line of the literature was added because once the graph is estimated, several machine learning methods are being used. Could you streamline this section by focusing on (1) whether it is difficult to forecast returns, (2) additional information is usually useful or not. Finally, how do the ML methods used in this paper add to the literature of returns forecasting (data is new, asynchronous, asymmetric, etc; methods are novels because they rely on two steps etc ... )}

 

\section{Data} \label{sec:data}
The stock data used in this study \mcc{cover} several of the world’s largest markets by market capitalization, including the New York Stock Exchange (NYSE), Nasdaq, the Shanghai Stock Exchange (SSE), and the Shenzhen Stock Exchange (SZSE). Daily U.S. stock data are \mcc{sourced} from the Center for Research in Security Prices\footnote{https://www.crsp.org/} (CRSP), while daily Chinese stock data are sourced from the Wind Database\footnote{https://www.wind.com.cn/}. The data \mcc{span} the period from 2014 through 2021. This selection of data enables us to investigate the transferability of signals across the world’s largest and most liquid equity markets \mcc{operating under non-overlapping trading sessions}.

In this paper, we rely on the market excess return of a stock, defined as the difference between the raw return of its price and the return of an exchange-traded fund (ETF) representing overall stock market performance. We use both pvCLCL returns and OPCL returns \mc{in one market} to forecast OPCL returns \mc{in the other market} (see Section~\ref{sec:def} for a more detailed justification of this setting). The pvCLCL logarithmic raw return for stock $i$ on day $t$ can be calculated by:  \MC{note I replaced stock $s$ by $i$ as it's more standard notation.}
\begin{equation}
    R_{i, \text{pvCLCL}}^{(t)} = \mcc{\log}\frac{p_{i,\text{cl}}^{(t)}}{p_{i,\text{cl}}^{(t-1)}},
\end{equation}
while the OPCL logarithmic raw return for stock $i$ on day $t$ can be calculated by:
\begin{equation}
    R_{i, \text{OPCL}}^{(t)} = \mcc{\log}\frac{p_{i,\text{cl}}^{(t)}}{p_{i,\text{op}}^{(t)}}. 
\end{equation}
Here $p_{i,\text{cl}}^{(t)}$ and $p_{i,\text{op}}^{(t)}$ denote the closing and opening price of stock $i$ on day $t$ respectively. Then the market excess return of stock $i$ on day $t$ can be defined as:
\begin{equation}
    r_{i}^{(t)} = R_{i, \text{pvCLCL}}^{(t)} - R_{\text{ETF}, \text{pvCLCL}}^{(t)}
\end{equation}
for pvCLCL returns, or
\begin{equation}
r_{i}^{(t)} = R_{i, \text{OPCL}}^{(t)} - R_{\text{ETF}, \text{OPCL}}^{(t)}
\end{equation}
for OPCL returns.
We use SPY as the market ETF in the U.S. and 513500.SH in China.

We select 500 stocks with the highest average market capitalizations over the years covered in the dataset from each country.\footnote{\mcc{This universe selection relies on full-sample information (average market capitalization over 2014--2021) and therefore introduces a mechanical look-ahead component. We adopt it as a pragmatic way to focus on continuously traded, highly liquid stocks and reduce missing observations. However, we caution that a fully investable design would require time-$t$ reconstitution based solely on lagged market capitalization information. Importantly, our main qualitative finding is directional asymmetry (the influence of the U.S. market on the Chinese market being stronger than the reverse effect), which is unlikely to be driven solely by this selection procedure. Nonetheless, we consider time-local universe formation as a valuable extension.}}
Unless otherwise specified, all returns mentioned in the following contents refer to market excess returns. \MC{this has some lookahead bias, if you average the market cap over the entire sample.. We can either not provide details here at all, and without explicitly saying "over the years covered in the data set -- or -- acknowledge and defend against (I added the paragraph below), as it's unlikely to change the main conclusion. What do you all think?"}

To mitigate the influence of extreme values and potential outliers, we \mcc{apply winsorization to the training-sample returns of each stock, replacing observations below the 0.5th percentile with the 0.5th percentile value and those above the 99.5th percentile with the 99.5th percentile value.}

\section{Methodology}
\label{sec:method}
This study aims to predict individual stock returns \mcc{using a cross-market directed bipartite graph}. The prediction framework consists of two main stages. First, we build a directed bipartite graph using \mcc{return data} from two markets within a look-back training window. This graph identifies \mcc{cross-market predictive links}: if a directed edge connects two stocks, the stock at the source of the edge \mcc{is treated as a predictor} for forecasting the returns of the stock at the destination. In the second stage, we apply various machine learning methods to forecast returns based on the identified relationships.

\subsection{Directed Bipartite Graph}
\label{sec:debigraph}
A graph can be defined as $\mathcal{G} = (\mathcal{V}, \mathcal{E})$, where $\mathcal{V}$ represents the vertex set and $\mathcal{E}$ represents the edge set. $\mathcal{G}$ is called bipartite if $\mathcal{V}$ can be divided into two disjoint sets $\mathcal{X}$ and $\mathcal{Y}$ such that all edges have one endpoint in $\mathcal{X}$ and another in $\mathcal{Y}$. \mcc{We denote a directed edge from $v_i$ to $v_j$ as $e_{ij}$ with associated weight $w_{ij}$.}
%
%
For a bipartite graph $\mathcal{G}$, the biadjacency matrix $\mathbf{B}$ is defined where rows correspond to nodes in $\mathcal{X}$, columns correspond to nodes in $\mathcal{Y}$, and each entry $b_{ij}$ contains the weight $w_{ij}$ of edge $e_{ij}$.

We \mc{represent} two different markets, the source market $\mathcal{X}$ and target market $\mathcal{Y}$, as two vertex sets, where stocks in each market \mc{are interpreted} as nodes. Edges originate from nodes in $\mathcal{X}$ and point to nodes in $\mathcal{Y}$. 
For a specific period of time $w$, which is the look-back training window in the experiment, the daily return vector of the $j\text{th}$ stock in market $\mathcal{X}$ is 
$$\bm{x} = [r_{X_j}^{(t-l-w)}, r_{X_j}^{(t-l-w+1)}, ..., r_{X_j}^{(t-l-1)}]^\intercal,$$ 
where $r_{X_j}^{(t)}$ is the return of the $j\text{th}$ stock on day $t$. The daily return vector of the $i\text{th}$ stock in market $\mathcal{Y}$ is $$\bm{y} = [r_{Y_i}^{(t-w)}, r_{Y_i}^{(t-w+1)}, ..., r_{Y_i}^{(t-1)}]^\intercal.$$
%
\mcc{The lag parameter $l$ captures the temporal ordering induced by non-overlapping trading sessions, ensuring that returns in the source market precede those in the target market.} 
Note that in our study the calculation with $t$ uses the trading calendar rather than the natural calendar.

\begin{figure}[htbp]
  \centering
  \includegraphics[width=0.5\linewidth]{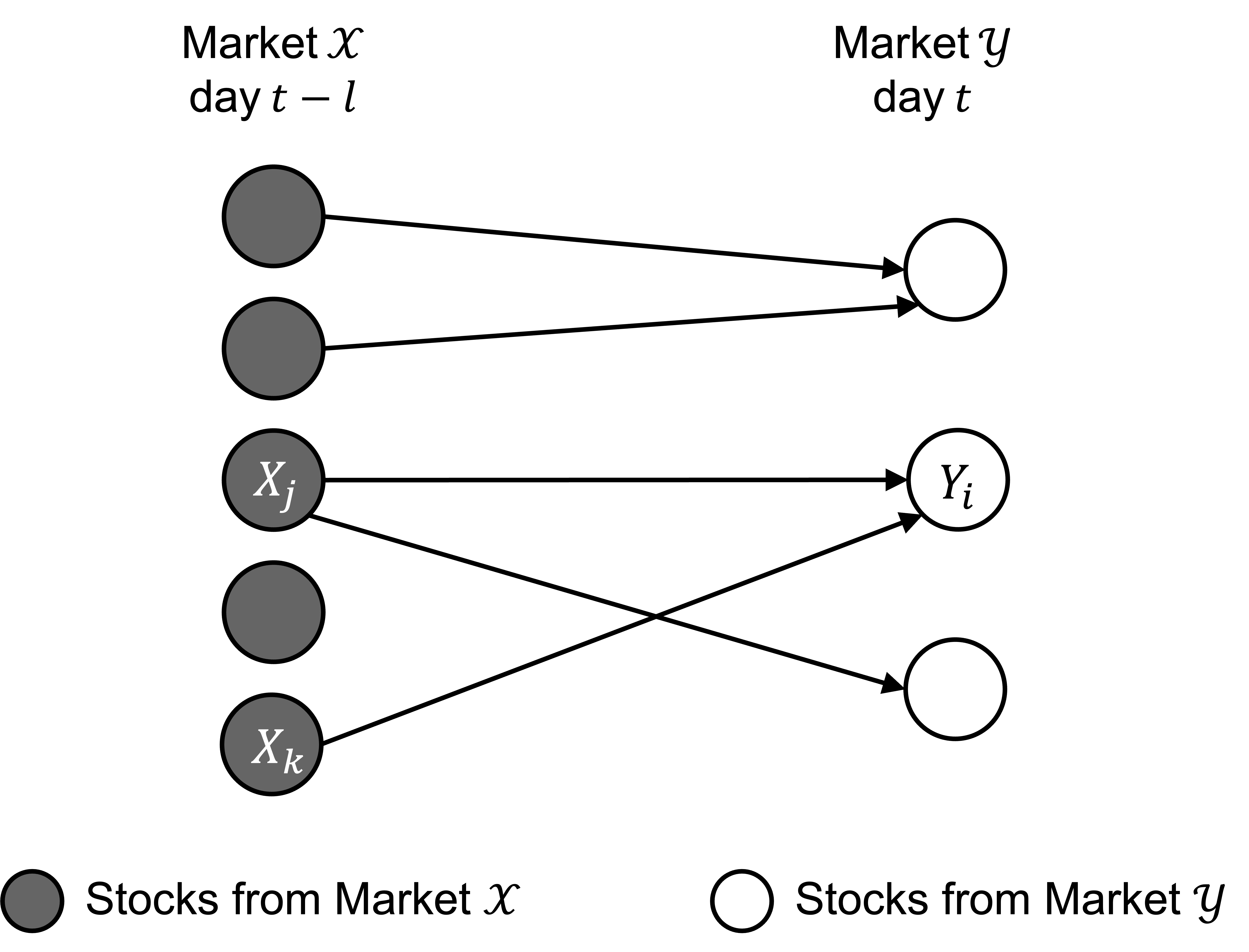}
  \caption{\mcc{Schematic illustration of the directed bipartite graph linking source-market stocks to target-market stocks based on significant predictive relationships.}}
  \label{fig:bpgraph_ex}
\end{figure}

\mcc{This time-ordered screening procedure induces a directed bipartite graph, where nodes in the source market $\mathcal{X}$ are connected to nodes in the target market $\mathcal{Y}$ whenever statistically significant predictive links are detected within the rolling training window. Figure~\ref{fig:bpgraph_ex} provides a schematic illustration of this bipartite structure.}

\MG{you could mode here the Figure: Example of a bipartite graph.}\MC{done, with option H to force in place + also added the sentence before the figure as up until this point, there has been no mention of a bipartite graph.}

\mcc{For each ordered pair $(X_j, Y_i)$, we estimate a univariate linear regression of $\bm{y}$ on $\bm{x}$ within the look-back window.}
We quantify such relationship using the t-statistic from regression, defined as
\begin{equation}
    t_{\beta} = \frac{\beta}{s_e/\sqrt{\sum_{i=1}^w{(x_i-\bar{x})^2}}}\label{eq1}.
\end{equation}
Here, $\beta$ is the slope coefficient of the simple linear regression, given by
$ \beta = \frac{\operatorname{cov}(\bm{x},\bm{y})}{\operatorname{var}(\bm{x})}$ 
and $s_e$ denotes the standard error of the regression
$    s_e = \sqrt{\frac{\text{SSE}}{w-2}}$. 
$\text{SSE}$ denotes the sum of squared  \mcc{residuals}, 
  $  \text{SSE} = \sum_{i=1}^w{(y_i - \hat {y_i})^2}$,
where
    $\hat{\bm{y}} = \beta \bm{x} + \bm{\alpha}$, and 
    $\bm{\alpha} = \bar{\bm{y}} - \beta \bar{\bm{x}}$.  
Here, $\hat {y_i}$ is the fitted value of $y_i$, and $\alpha$ is the intercept of the regression line. 
\MG{Is univariate testing a common procedure? Would a multiple hypothesis testing across hundreds of stocks be possible? Acknowledge the pros/cons.} \MC{I added the paragraph below and cited Fan/Harvey as new refs}

\mcc{The use of pairwise univariate screening serves primarily as a computationally tractable sparsification device rather than as a formal structural inference procedure. Similar marginal screening approaches are common in high-dimensional predictive settings where the objective is feature selection rather than causal identification (see, e.g., \citet{fan2008_sure_screening}; \citet{hastie2009elements}). We recognize that testing across a large number of stock pairs raises multiple-testing considerations and may introduce spurious edges in finite samples (cf. \citet{Harvey2015_multiple_testing}). In principle, false discovery rate or multiple-comparison corrections could be applied. However, our primary goal is to construct a predictive graph that enhances out-of-sample forecasting performance rather than to perform statistical inference on individual edges. We therefore treat the screening step as a model-selection heuristic and assess its validity through out-of-sample forecasting performance and robustness analyses.}

\begin{figure}[htbp]
  \centering
  \includegraphics[width=0.65\linewidth]{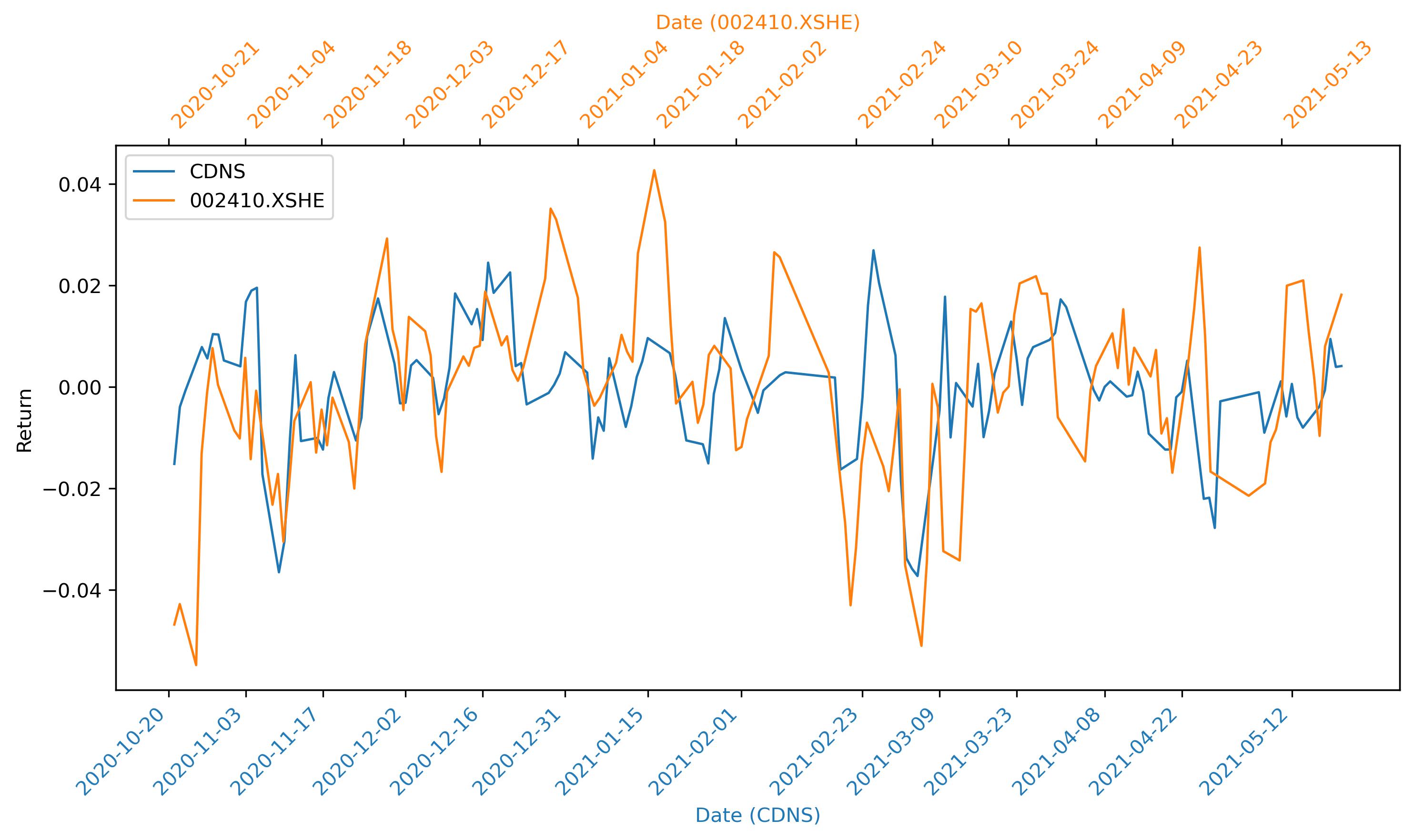}
  \caption{\mcc{Example time series of U.S. pvCLCL returns for CDNS and Chinese OPCL returns for 002410.XSHE over the rolling training window. The series are shown for illustrative purposes to highlight cross-market co-movement underlying the detected predictive link.}}
  \label{fig:ret_in_250_4}
\end{figure}

Figure~\ref{fig:ret_in_250_4} shows the return time series for an example pair of U.S. and Chinese technology stocks, CDNS (pvCLCL returns) and 002410.XSHE (OPCL returns), smoothed with a three-day moving average for visualization purposes. Here $l=1$ and $w=250$. The t-statistic from the regression of 002410.XSHE on CDNS is high during the period shown, \mcc{illustrating a statistically significant cross-market predictive relation within the training window under the linear screening specification.}

In our setting, either the U.S. or the Chinese market can be treated as \mc{the source} market $\mathcal{X}$, with the target market of prediction serving as market $\mathcal{Y}$. After performing the regression t-test above for all \mcc{ordered stock pairs} in market $\mathcal{X}$ and $\mathcal{Y}$, we set a threshold to filter the resulting t-statistics by magnitude. 
\mc{We introduce an explicit threshold parameter, denoted by $\tau$, to facilitate later reference. In our experiments, we set $\tau = 2$, and select edges whenever $\mcc{|t_{\beta}| > \tau}$, corresponding approximately to conventional significance levels under standard asymptotic approximations.}
If the magnitude of the t-statistics for $\bm{x}$ and $\bm{y}$ is larger than $\tau$, we select the return of $X_j$ on day $t-l$ to predict the return of $Y_i$ on day $t$. This selection forms a \mc{directed} edge \mc{in the graph} pointing from $X_j$ to $Y_i$.  
\mc{Note that the thresholding step is used purely as a sparsification mechanism, aimed at denoising the signal and improving computational tractability rather than constituting a formal multiple-testing correction.}
A sample directed bipartite graph is shown in Figure~\ref{fig:bpgraph_ex}, where $X_j$ and $X_k$ on day $t-l$ are selected to predict $Y_i$ on day $t$. 
\mc{This construction yields a time-lagged cross-market predictive network that can be naturally interpreted as a directed bipartite graph.}

\begin{figure}[htbp]
  \centering
  \includegraphics[width=0.9\linewidth]{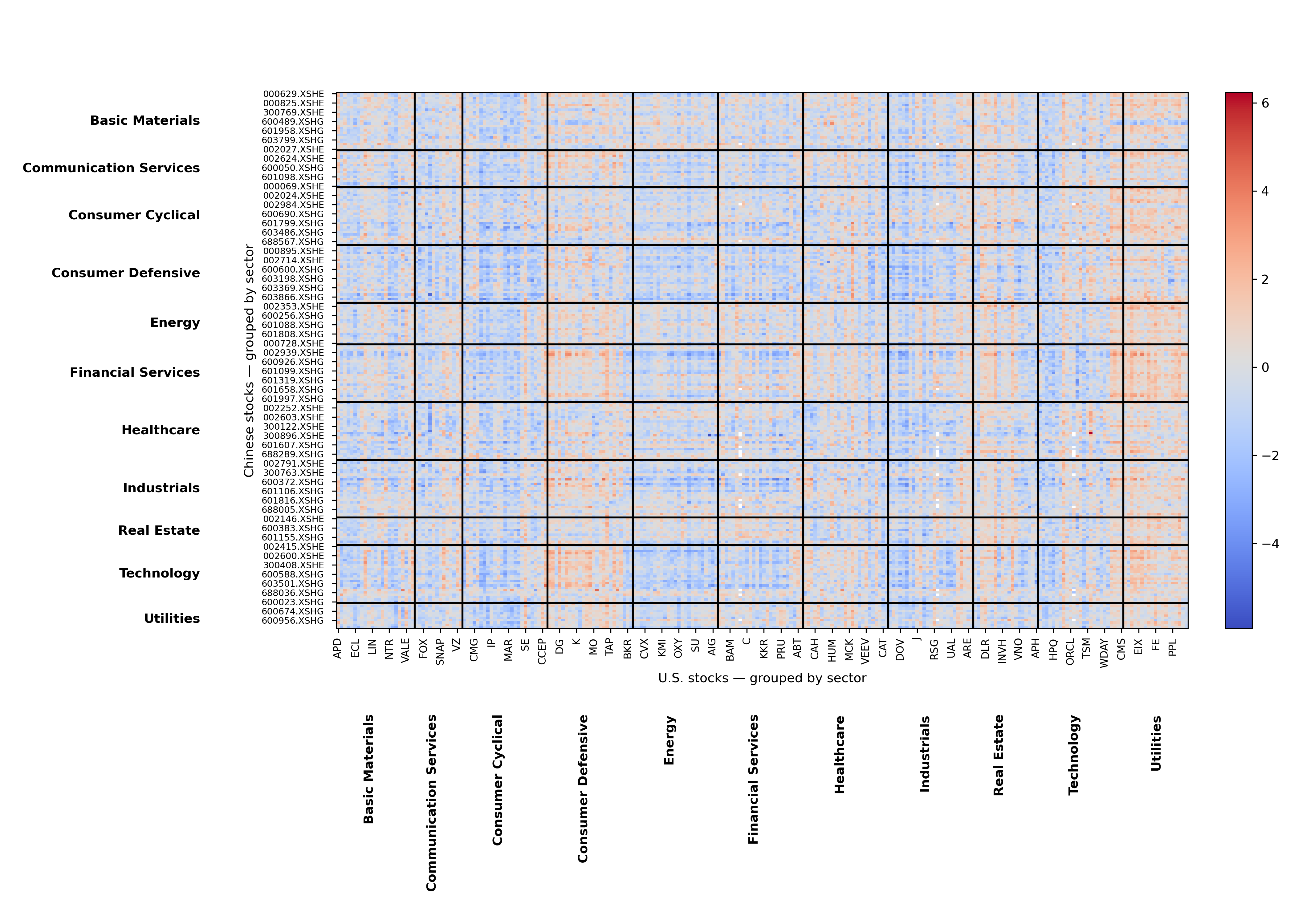}
  \vspace{-5mm}
\caption{\mcc{Heatmap of the directed biadjacency matrix for a representative trading day. Rows correspond to Chinese stocks and columns to U.S. stocks, grouped by sector. Each entry represents the t-statistic from the rolling-window regression of Chinese returns on lagged U.S. returns. Colour intensity reflects the magnitude and sign of the predictive relationship.}}
 \label{fig:adjmat_heatmap}
\end{figure}

Figure~\ref{fig:adjmat_heatmap} presents a section of the heatmap corresponding to the biadjacency matrix of the U.S.–Chinese stock network on 21 October 2021. To illustrate the structure more clearly, we select 25 representative stocks from each sector. For sectors containing fewer than 25 stocks in the original dataset, all available stocks are included, resulting in \mcc{254 U.S. stocks and 235 Chinese stocks in this visualization}. Each row corresponds to a Chinese stock, while each column represents a U.S. stock. The colour intensity represents the value of the t-statistic, \mcc{and black grid lines delineate sectoral boundaries}. 
This \mc{visualization} shows that cross-market \mcc{predictive connectivity} is not restricted to \mc{within-sector interactions} (which would lead to a block-diagonal structure), \mcc{thereby motivating a flexible cross-market predictive framework}.

\begin{figure}[htbp]
  \centering
    \includegraphics[width=0.6\linewidth]{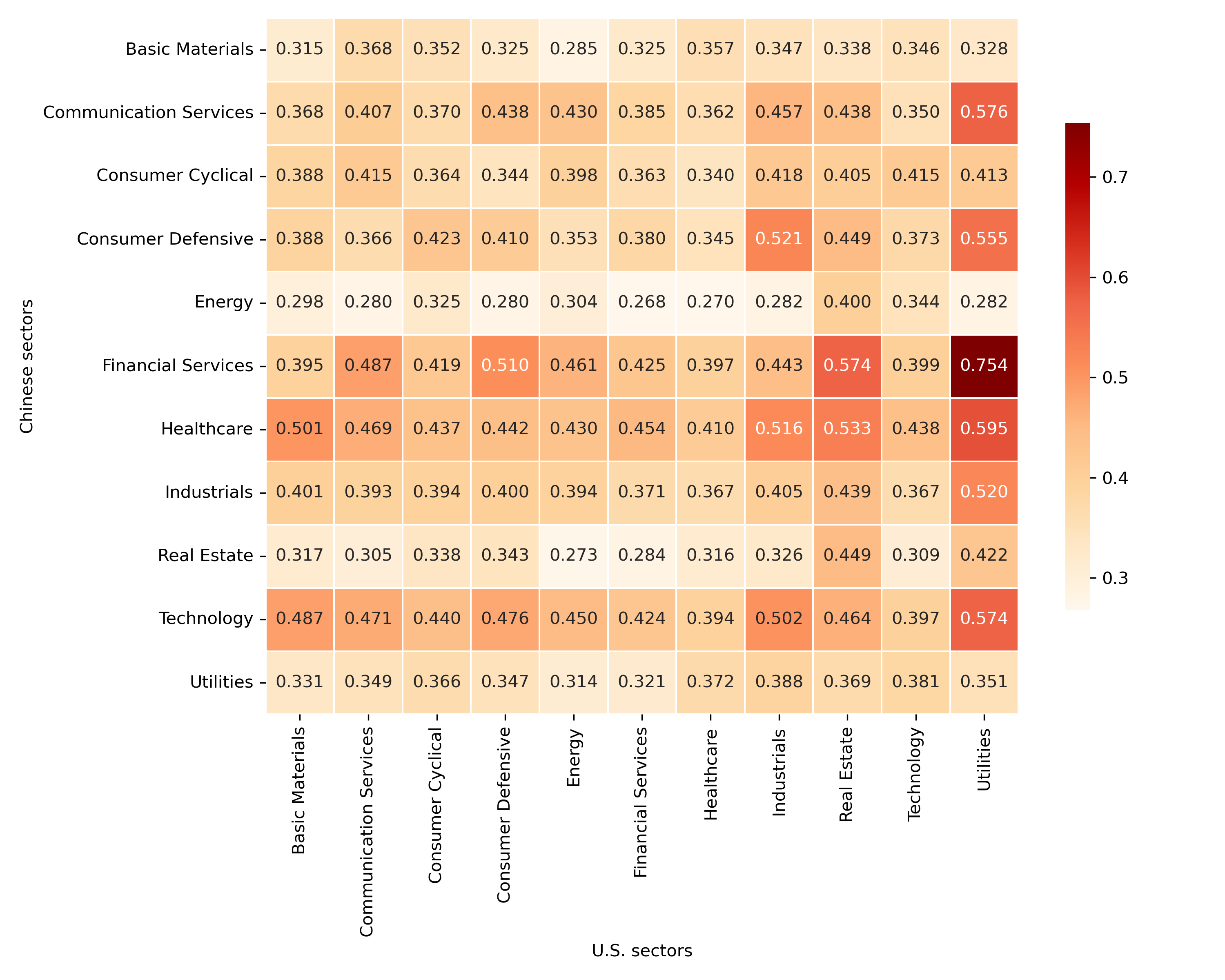}
\vspace{-3mm}    
    \caption{\mcc{Sector-level heatmap of the absolute median t-statistic in the time-averaged biadjacency matrix of the directed cross-market graph. Rows correspond to Chinese sectors and columns to U.S. sectors. Each entry reports the median absolute predictive strength across all stock pairs within the corresponding sector-by-sector block.}}
    \label{fig:adjmat_heatmap_sec_abs}
\end{figure}

\mcc{To summarize cross-market structure over time, we average the daily biadjacency matrices across the full sample period, obtaining an aggregate representation of predictive linkages.}  
We then compute, for this time-averaged matrix, the median of absolute t-statistics within each sector-by-sector block of the corresponding heatmap (Figure~\ref{fig:adjmat_heatmap_sec_abs}).  
\mcc{This aggregation highlights systematic sectoral dependencies rather than stock-specific effects.} For example, the financial services sector in the Chinese market exhibits strong predictive links with the utilities sector in the U.S. market.

\begin{figure}[t]
  \centering
  \includegraphics[width=0.9\linewidth]{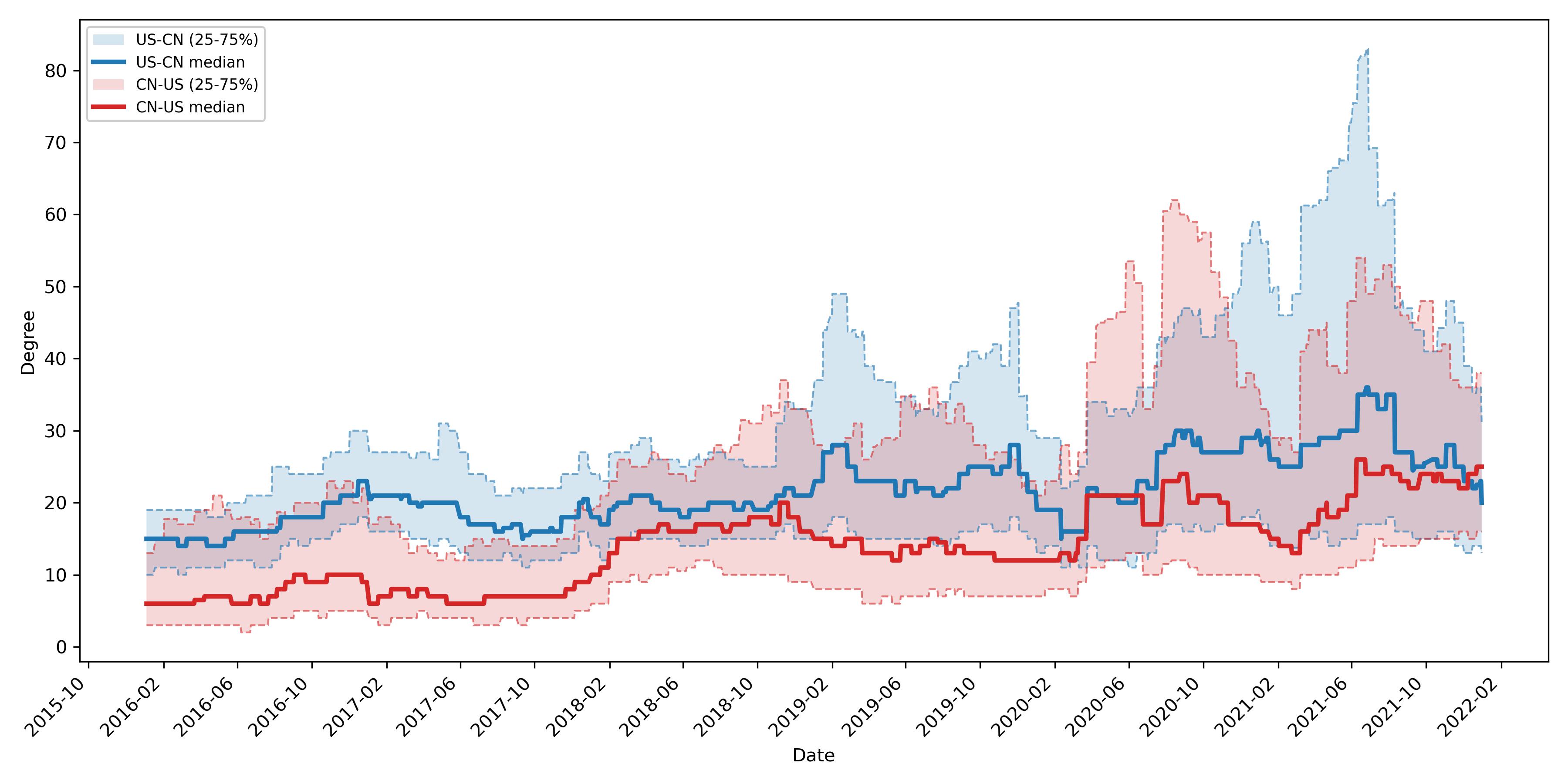}
  \caption{\mcc{The figure shows the 25th, 50th, and 75th percentiles of the in-degree distribution of target nodes by day. US-CN represents the number of U.S. pvCLCL nodes selected to predict Chinese OPCL returns, while CN-US represents the number of Chinese pvCLCL nodes selected to predict U.S. OPCL returns.} 
  }
  \label{fig:avgdeg}
\end{figure}

We also examine the cross-market relations over time. 
Figure~\ref{fig:avgdeg} shows how the in-degree of all nodes in set $\mathcal{Y}$ \mc{evolves} over time. For each day, the 25th, 50th, and 75th percentiles of the in-degree distribution are computed across all target nodes. The blue curves represent the number of U.S. pvCLCL nodes selected to predict Chinese OPCL returns, \mc{while} the red curves \mc{correspond to} the number of Chinese pvCLCL nodes selected to predict U.S. OPCL returns. As time progresses, the \mcc{in-degree in both directions increases}, suggesting strengthening \mcc{cross-market predictive connectivity over the sample period}.

\subsection{Predictive Analysis with Machine Learning}
\label{sec:def}

\begin{figure}[htbp]
  \centering
  \includegraphics[width=\linewidth]{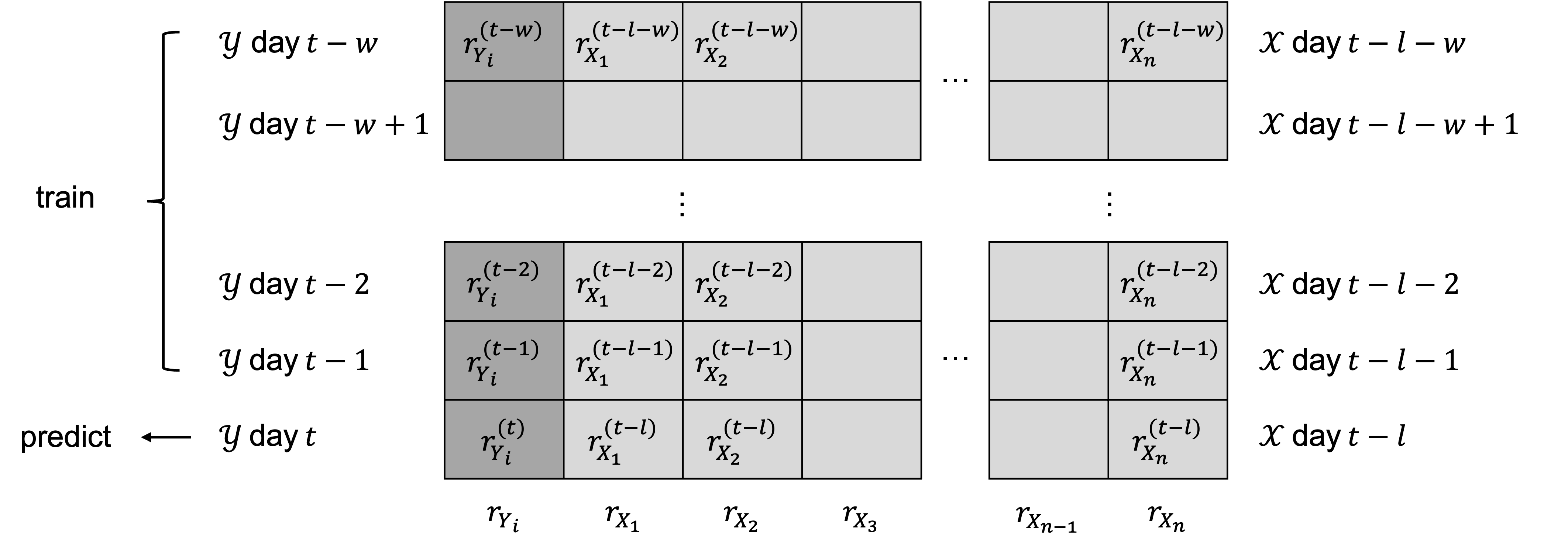}
\caption{\mcc{Schematic illustration of the rolling training and prediction framework. For each target stock $Y_i$, returns over the look-back window $[t-w, t-1]$ are regressed on lagged source-market returns over $[t-l-w, t-l-1]$. The bottom row represents the out-of-sample prediction of $r_{Y_i}^{(t)}$ using source returns observed at $t-l$, thereby preserving temporal ordering and eliminating look-ahead bias.}}
\label{fig:reg_data}
\end{figure}

In order to predict \mc{the} return of stock $Y_i$ on day $t$, we use training data from day $t-w$ to day $t-1$ for market $\mathcal{Y}$ and from day $t-l-w$ to day $t-l-1$ for market $\mathcal{X}$. 
Since we wish to predict $r_{Y_i}^{(t)}$ by \mc{using information} from market $\mathcal{X}$, we select $n$ stocks, i.e., $X_1, X_2, ..., X_n$ from market $\mathcal{X}$, 
\mc{corresponding to those stocks that exhibit the strongest cross-market predictive associations} with $Y_i$ according to the t-statistic defined above. 
%
Their daily returns on day $t-l$ are $r_{X_1}^{(t-l)}, r_{X_2}^{(t-l)}, ..., r_{X_n}^{(t-l)}$. The data used for training and prediction are illustrated in Figure~\ref{fig:reg_data}. 
\mcc{All predictor selection is performed within the rolling training window to avoid look-ahead bias.}

\begin{figure}[htbp]
  \centering
  \includegraphics[width=0.6\linewidth]{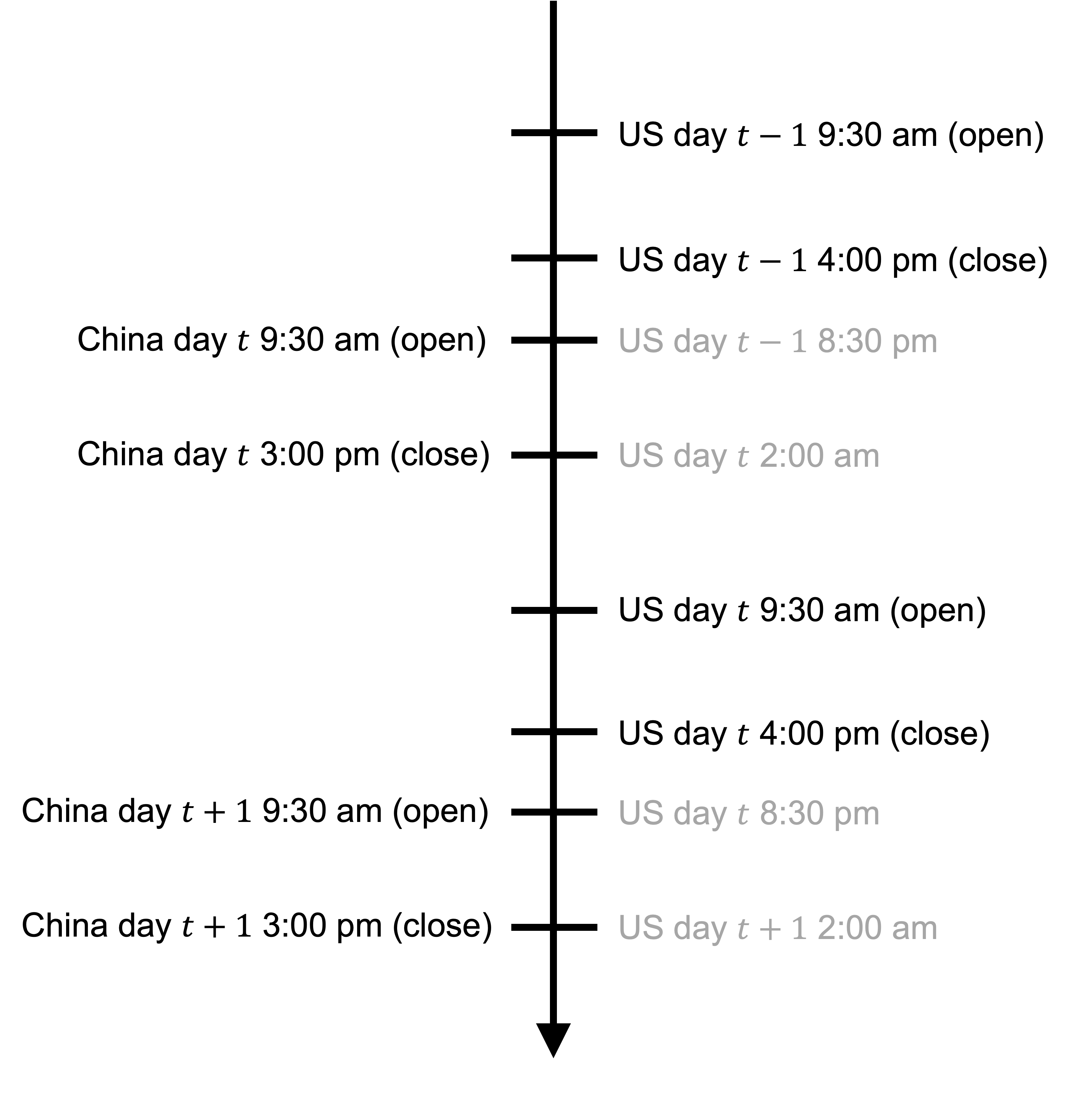}
  \caption{\mcc{Timeline of opening and closing times for the U.S. and Chinese stock markets. The non-overlapping trading sessions induce a natural temporal ordering of information, with U.S. day $t-1$ close preceding Chinese day $t$ trading, and Chinese day $t$ close preceding U.S. day $t$ trading.}}
  \label{fig:timeline}
\end{figure}

The U.S. market is open from 9:30am to 4:00pm \mcc{U.S. Eastern Time (ET)}, while the Chinese market is open from 9:30am to 11:30am, and 1:00pm to 3:00pm \mcc{China Standard Time (UTC+8)}. There is no overlap between the two trading periods, as shown in the time zone diagram in Figure~\ref{fig:timeline}, under the standard time difference. Note that adjusting for daylight saving time does not result in any overlap between the trading sessions. We predict OPCL returns for both countries. We set $l=1$ when predicting Chinese stocks \mc{using} the latest information from the U.S. market, and $l=0$ \mc{in the reverse direction}. 
\mcc{This timing structure ensures that predictor information from the source market is fully observable prior to the opening of the target market.}

We build the forecasting model as follows:
\begin{equation}
    r_{Y_i}^{(t)}=F_i(r_{X_1}^{(t-l)}, r_{X_2}^{(t-l)}, ..., r_{X_n}^{(t-l)};\theta)+\epsilon_i^{(t)}.
\end{equation}
Here, the function $F_i$ represents the different machine learning methods we use, and $\theta$ refers to the parameters that are estimated for each machine learning model. The aim is to identify a model that can generate \mc{accurate out-of-sample predictions of} $r_{Y_i}^{(t)}$ so that a high SR can be achieved.

We applied a total of ten machine learning models to forecast returns. They include: Ordinary Least Squares (OLS), Least Absolute Shrinkage and Selection Operator (LASSO), Ridge Regression (RIDGE), Support Vector Machine (SVM), Extreme Gradient Boosting (XGBoost), Light Gradient Boosting Machine (LGBM), Random Forests (RF), Adaptive Boosting (AdaBoost), ensemble by results average (ensemble-avg) and ensemble by results median (ensemble-med). \mcc{This range of models spans linear, regularized, kernel-based, tree-based, and ensemble approaches, allowing us to assess whether cross-market predictive gains depend on model class or are robust across specifications.} We describe each model in detail below. \mcc{All models are estimated within each rolling training window and evaluated out-of-sample to ensure temporal validity.}

\begin{itemize}
\item \textbf{Ordinary Least Squares (OLS)}: 
The main idea of OLS is to estimate regression coefficients by choosing parameter values that minimize the sum of squared residuals between observed and predicted values. Specifically, the model is defined as:
\begin{equation}
    r_{Y_i}^{(t)}=\alpha_i+\sum_{j=1}^{n}\beta_{ij} r_{X_j}^{(t-l)}+\epsilon_i^{(t)}.
\end{equation}
The linear model is fit with an objective of minimizing the residual sum of squares (RSS):
\begin{equation}
    \min_{\alpha_i, \bm{w}} \left\| \bm{y} - \alpha_i \bm{1} - \mathbf{X}\bm{w} \right\|_2^2 .
\end{equation}
$\bm{y} \in \mathbb{R}^{d}$ is the vector of returns \mc{corresponding to} $r_{Y_i}^{(t)}$, $\mathbf{X} \in \mathbb{R}^{d \times n}$ is the matrix of predictors where each row is $[r_{X_1}^{(t-l)}, \dots, r_{X_n}^{(t-l)}]$, and $\bm{w} = [\beta_{i1}, \dots, \beta_{in}]^\top$ is the \mc{associated} coefficient vector. Here $d=w$, which is the length of training window, i.e., the number of time points.

\item \textbf{Least Absolute Shrinkage and Selection Operator (LASSO)}:
The OLS method often leads to low bias but high variance~\citep{hastie2009elements}. Shrinkage methods are introduced to mitigate this problem, and LASSO is one of them. It uses $\ell_1$-norm regularization to impose a penalty on the size of regression coefficients~\citep{hastie2009elements}. The objective function is given by:

\begin{equation}
    \min_{\alpha_i, \bm{w}} \left( \frac{1}{2d} \left\| \bm{y} - \alpha_i \bm{1} - \mathbf{X}\bm{w} \right\|_2^2 + \lambda \|\bm{w}\|_1 \right).
\end{equation}
Here $\lambda$ is the regularization parameter.

\item \textbf{Ridge Regression (RIDGE)}:
RIDGE is another type of shrinkage method. It adds $\ell_2$-norm regularization to the linear least squares loss function. The objective function is given by:
\begin{equation}
    \min_{\alpha_i, \bm{w}} \left( \left\| \bm{y} - \alpha_i \bm{1} - \mathbf{X}\bm{w} \right\|_2^2 + \lambda \|\bm{w}\|_2^2 \right).
\end{equation}

\item \textbf{Support Vector Machine (SVM)}:
SVMs can tackle complex learning problems \mc{while retaining} the analytical simplicity of linear models. With kernel functions, this method avoids direct computation in high-dimensional spaces, \mc{enabling} nonlinear learning using a linear algorithm in the feature space~\citep{hearst1998support}. We use the radial basis function kernel \mc{throughout} our experiment. The goal is to minimize the following \mcc{dual optimization problem} with respect to the Lagrange multipliers:
\begin{equation}
\begin{aligned}
    \min_{\bm{\alpha}} & \frac{1}{2} \sum_{j=1}^{n} \sum_{k=1}^{n} \alpha_j \alpha_k y_j y_k {K(\bm{x}_j, \bm{x}_k)} - \sum_{j=1}^{n} \alpha_j \\
    \text{s.t.} \quad & \sum_{j=1}^{n} \alpha_j y_j = 0 \\
    & 0 \le \alpha_j \le C, \quad j=1, 2, \dots, n .
\end{aligned}
\end{equation}
Here $\alpha_j$ is the Lagrange multiplier, $C$ is a hyperparameter that controls the trade-off between the flatness of the function and the amount \mc{by} which deviations larger than $\epsilon$ are tolerated,  $K(\bm{x}_j, \bm{x}_k)$ is the kernel function, $y_j$ is $r_{Y_i}^{(t-w+j-1)}$ in the training window, and $\bm{x}_j$ is its corresponding vector of predictors $[r_{X_1}^{(t-w+j-1-l)}, \dots, r_{X_n}^{(t-w+j-1-l)}]^\intercal$.

\item \textbf{Extreme Gradient Boosting (XGBoost)}:
XGBoost is a scalable end-to-end tree boosting method~\citep{chen2016xgboost}. It implements parallel and distributed computing \mcc{to accelerate training}. The model is \mc{defined by the following} equation:
\begin{equation}
    \hat{y}_j = \phi(\bm{x}_j) = \sum_{k} f_k(\bm{x}_j), \quad f_k \in \mathcal{F},
\end{equation}
where $\mathcal{F}$ is the space of regression trees, and $f_k$ is one independent tree. The objective function is given by:
\begin{equation}
    \min {\sum_{j} l(\hat{y}_j, y_j) + \sum_{k} \Omega(f_k)}, 
\end{equation}
where
\begin{equation}
    \Omega(f) = \gamma M + \frac{1}{2}\lambda\|w\|^2.
\end{equation}
Here $l$ is a differentiable convex loss function, $\Omega(f)$ is the regularization term, $M$ is the number of leaves, $w$ is the leaf weight, and $\gamma$ and $\lambda$ are the corresponding regularization parameters.

\item \textbf{Light Gradient Boosting Machine (LGBM)}:
LGBM is another gradient boosting method that \mc{improves computational efficiency compared with standard} gradient boosting tree algorithms~\citep{ke2017lightgbm}. Two key techniques \mc{employed by LGBM} are Gradient-Based One-Side Sampling and Exclusive Feature Bundling. The former retains instances with large gradients and randomly samples those with small gradients. The latter combines mutually exclusive sparse features, which never take nonzero values at the same time, into a single combined feature, effectively reducing computational complexity~\citep{ke2017lightgbm}.

\item \textbf{Random Forests (RF)}:
Random forests consist of an ensemble of decision trees. At each node of each tree, the algorithm randomly selects a subset of features to consider for splitting. Each tree is grown using bootstrap sampling of the data~\citep{breiman2001random}. \mc{By combining these de-correlated trees, the final prediction is obtained by averaging their outputs~\citep{hastie2009elements}.}

\item \textbf{Adaptive Boosting (AdaBoost)}:
AdaBoost combines multiple weak learners to form a strong \mc{learner}. Each weaker \mc{learner} is trained to correct the errors made by the previous \mc{learners}. The algorithm \mcc{iteratively reweights training observations based on their absolute prediction errors}, so that more emphasis is given to instances with \mc{larger} errors from earlier learners. \mc{The final prediction is obtained by aggregating the weak learners}, summing their probabilistic predictions~\citep{freund1997decision}. We choose a decision tree regressor as the base learner in our experiment.

\item \textbf{Ensemble by results average (ensemble-avg)}:
For each stock and each day, we take the average of the prediction results from the eight methods above as the final output.

\item \textbf{Ensemble by results median (ensemble-med)}:
Similar to ensemble-avg, we take the median of the prediction results from the eight methods as the final output for each stock on each day.

\end{itemize}

\section{Experiments}
\label{sec:exp}
\mc{In this section, we conduct an extensive set of experiments to evaluate the cross-market predictability of individual stock returns and examine the economic relevance of the proposed graph-based framework.} 
\mcc{All results are obtained using a rolling-window estimation scheme and evaluated strictly out-of-sample.}

\subsection{Evaluation Metrics}
We use Profit and Loss (PnL) and Sharpe Ratio (SR) to evaluate the performance of forecasting methods. \mcc{We abstract from liquidity-optimized portfolio construction and explicit transaction cost modeling, and therefore interpret reported SRs as pre-cost measures of predictive strength rather than implementable performance.}

\begin{itemize}
\item \textbf{Profit and Loss (PnL)}:
The PnL on day $t$ is calculated with the following equation:
\begin{equation}
    PnL^{(t)} = \sum_i \text{sign}(s_i^{(t)}) \cdot r_i^{(t)} \cdot b_i^{(t)}. 
\end{equation}
Here $s_i^{(t)}$ denotes the predicted return of stock $i$ on day $t$, and $r_i^{(t)}$ denotes the actual return of stock $i$ on day $t$. \mc{$b_i^{(t)} = \min(0.001 \times mdv_{i}^{(21)}, L)$ is the amount of capital deployed on stock $i$, where $mdv_{i}^{(21)}$ denotes the median daily traded volume of stock $i$ over the 21-day interval preceding day $t$, and $L$ is the maximum limit to the bid. The parameter $L$ controls the maximum position size.} 
\mcc{This position-sizing rule serves as a coarse liquidity proxy, limiting exposure in less actively traded names.} 
Throughout our experiment, \mcc{$L$ is set to 100{,}000 USD for the U.S. market prediction and 1{,}500{,}000 CNY for the Chinese market prediction.}

\item \textbf{Sharpe Ratio (SR)}:
After computing daily PnLs of all stocks, we calculate the mean and standard deviation \mc{of the daily} PnL vector with length $T$, denoted as $\mu_{PnL}^{(T)}$ and $\sigma_{PnL}^{(T)}$, where $T$ is the length of predicting period in our experiment. 
\mc{The annualized SR is given by:}
\begin{equation}
    SR = \frac{\mu_{PnL}^{(T)}}{\sigma_{PnL}^{(T)}} \cdot \sqrt{252}.
\end{equation}
Here the scaling accounts for the fact that there are 252 trading days in a calendar year and annualizes \mcc{daily PnL variability}.
\end{itemize}

\mcc{Several practical limitations should be noted. First, the graph is obtained via large-scale pairwise screening and therefore may include spurious edges in the presence of multiple testing. 
\mcc{Second, our economic evaluation abstracts from transaction costs, market impact, short-sale constraints, and other trading frictions, so reported SRs reflect pre-cost predictive performance.}
Third, we do not implement liquidity-weighted portfolio construction or dynamic capacity controls; position sizes are capped but not optimized with respect to market depth. Consequently, the trading design is stylized rather than fully implementable. As emphasized by~\citet{cartea2025liquidity}, ignoring stock-level capacity constraints can substantially overstate the implementable value of predictive strategies. Our objective is to isolate and quantify directional cross-market predictive asymmetries rather than to construct a production-ready trading strategy.}

\subsection{Experimental Setup}
\label{sec:main_exp}

We use a 250-day training window and update \mc{both} the graphs and \mc{the predictive} models every 10 days. Prediction begins on the first trading day of 2016 and ends on the last trading day of 2021. \mcc{All models are re-estimated using a rolling-window scheme to ensure strict out-of-sample evaluation and avoid look-ahead bias.}

\subsubsection{Graph-Based Cross-Market Prediction}
\begin{itemize}
\item \textbf{Predicting the Chinese Market with the U.S. market}:
We let market $\mathcal{X}$ denote the U.S. market, and market $\mathcal{Y}$ denote the Chinese market. We use the most recent available U.S. returns as predictors to forecast Chinese returns, i.e., $l=1$, \mcc{reflecting the non-overlapping trading sessions and the temporal ordering of information flow}.

\item \textbf{Predicting the U.S. Market with the Chinese market}:
We let market $\mathcal{X}$ denote the Chinese market, and market $\mathcal{Y}$ denote the U.S. market. We also use the most recent available returns for forecasting, i.e., $l=0$, \mcc{since Chinese trading concludes before the U.S. market opens on the same calendar day}.
\end{itemize}

\subsubsection{Baseline}
\begin{itemize}
\item \textbf{Non-Graph-Based Same-Market Baseline}:
For each target stock, the previous 25 days of daily return data are used as predictive features. \mcc{The training window remains 250 days and models are updated every 10 days to ensure comparability with the graph-based specifications.}
This baseline model can be described with the following equation:
\begin{equation}
    r_{Y_i}^{(t)}=F_i(r_{Y_i}^{(t-25)}, r_{Y_i}^{(t-24)}, ..., r_{Y_i}^{(t-1)};\theta)+\epsilon_i^{(t)}.
\end{equation}
Note that the predictive features $r_{Y_i}^{(t-25)}, r_{Y_i}^{(t-24)}, ..., r_{Y_i}^{(t-1)}$ can be either all pvCLCL returns or all OPCL returns, while $r_{Y_i}^{(t)}$ is an OPCL return. 

\item \textbf{Graph-Based Same-Market Baseline}: 
Based on the methodology described in Section~\ref{sec:debigraph}, this baseline sets markets $\mathcal{X}$ and $\mathcal{Y}$ identical, so that for each stock its predictors are drawn from the same market. The return values of predictors are one-day ahead of the response values, i.e., $l=1$. \mcc{This specification isolates the incremental contribution of cross-market information relative to graph-based modeling per se.}
\end{itemize}

\subsection{Main Results}
\MC{Before, we also had this sentence (citing Castro 2022) at the start of this subsection; not sure if it's really needed to cite something that is so standard in the literature, but we could? \mc{Quantiles have often been used in decision-making in social activities such as banking and investment~\citep{castro2022portfolio}.}}

\mcc{We begin by evaluating the economic performance of the cross-market forecasting framework. Portfolio sorts based on model-implied signals are standard in the return predictability literature. For each day $t$, stocks are ranked by the absolute value of their predicted returns, $|\hat r_i^{(t)}|$. To examine how performance varies with signal strength, we construct six nested \mcc{quantile portfolios}:
\begin{itemize}
\item quantile 1 (qr1): all stocks;
\item quantile 2 (qr2): top 80\% of stocks ranked by $|\hat r_i^{(t)}|$;
\item quantile 3 (qr3): top 60\%;
\item quantile 4 (qr4): top 40\%;
\item quantile 5 (qr5): top 20\%;
\item quantile 6 (qr6): top 10\%.
\end{itemize}
These portfolios are nested, so that $\text{qr6} \subset \text{qr5} \subset \text{qr4} \subset \text{qr3} \subset \text{qr2} \subset \text{qr1}$. This construction allows us to assess whether stronger model signals translate into improved risk-adjusted performance. 
Importantly, the ranking at day $t$ is based solely on model predictions available at that date and does not use realized returns\mcc{, thereby avoiding look-ahead bias in portfolio formation}. \mcc{We first document that cross-market information and graph-based modeling contribute to improved forecasting performance.}}

\begin{figure}[htbp]
\centering
\begin{subfigure}{0.48\textwidth}
    \centering
    \includegraphics[width=\linewidth]{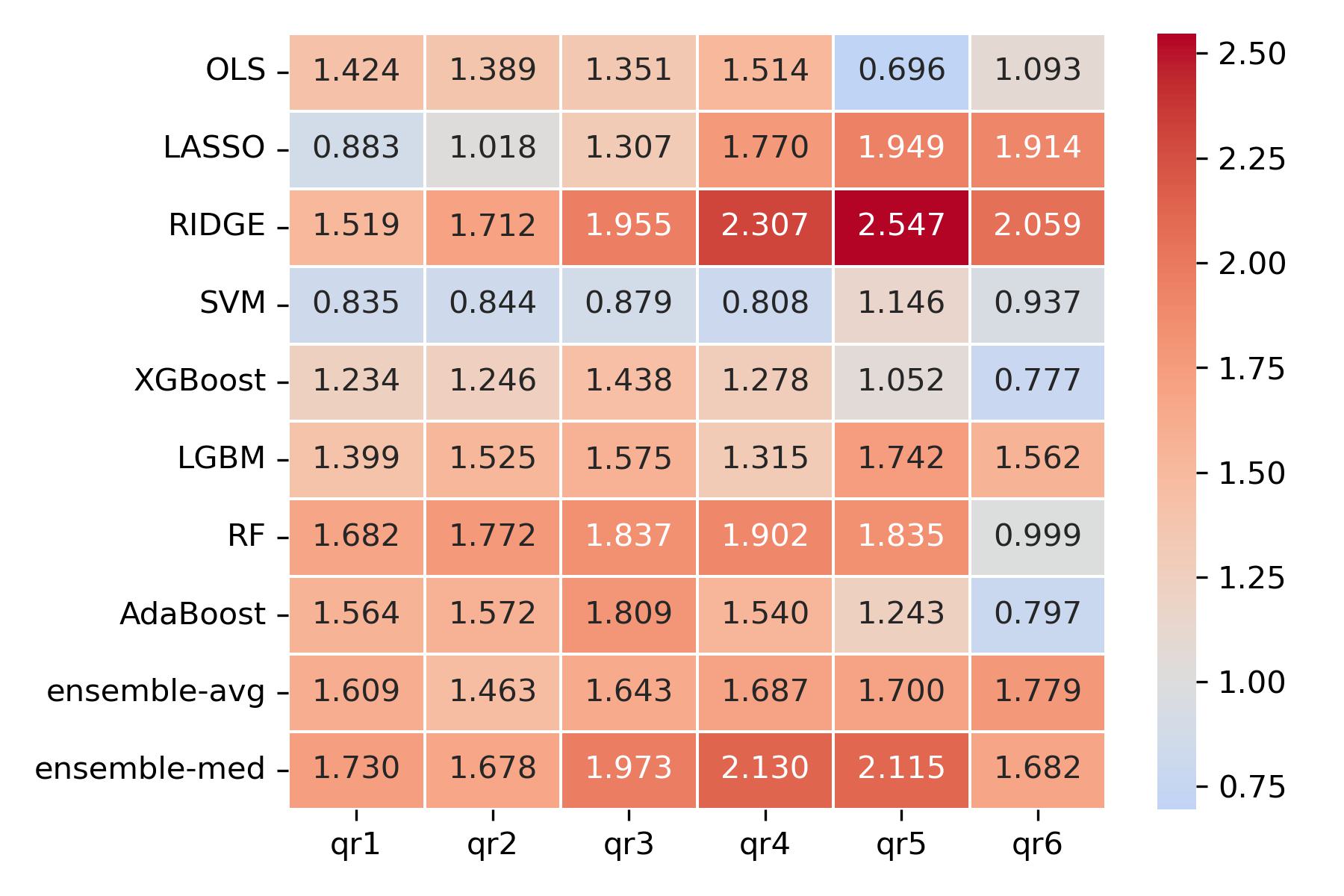}
        \caption{\mcc{Predictors: U.S. pvCLCL returns.}}
    \label{fig:US-CH_pvCLCL-OPCL_pred}
\end{subfigure}
\hfill 
\begin{subfigure}{0.48\textwidth}
    \centering
    \includegraphics[width=\linewidth]{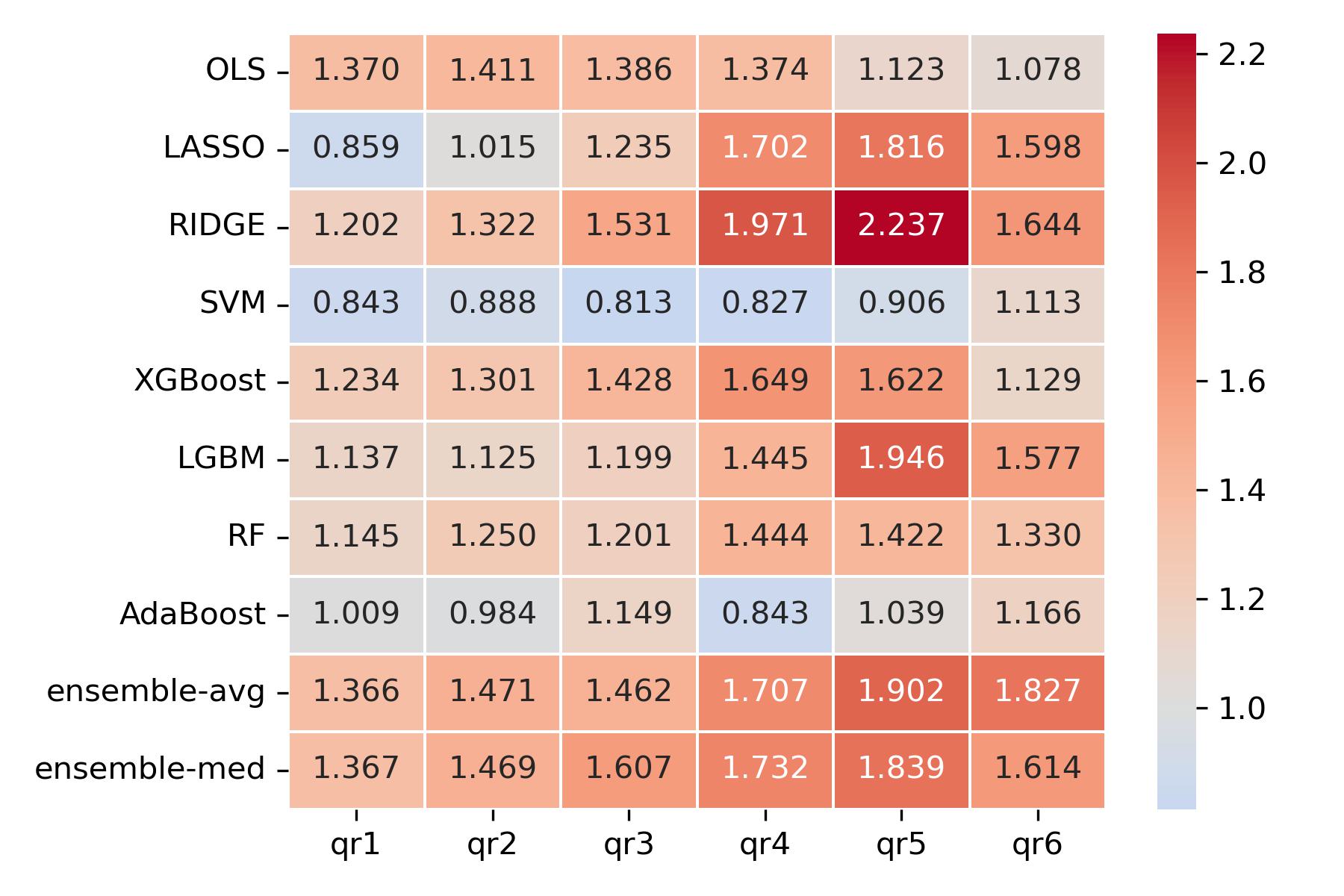}
    \caption{\mcc{Predictors: U.S. OPCL returns.}}
    \label{fig:US-CH_OPCL-OPCL_pred}
\end{subfigure}
\caption{\mcc{Sharpe Ratios for forecasting Chinese OPCL returns using U.S. pvCLCL and OPCL returns as predictors.}}
\label{fig:US-CH_pred}
\end{figure}
%

\begin{figure}[htbp]
    \centering
    \begin{subfigure}{0.48\textwidth}
        \centering
        \includegraphics[width=\linewidth]{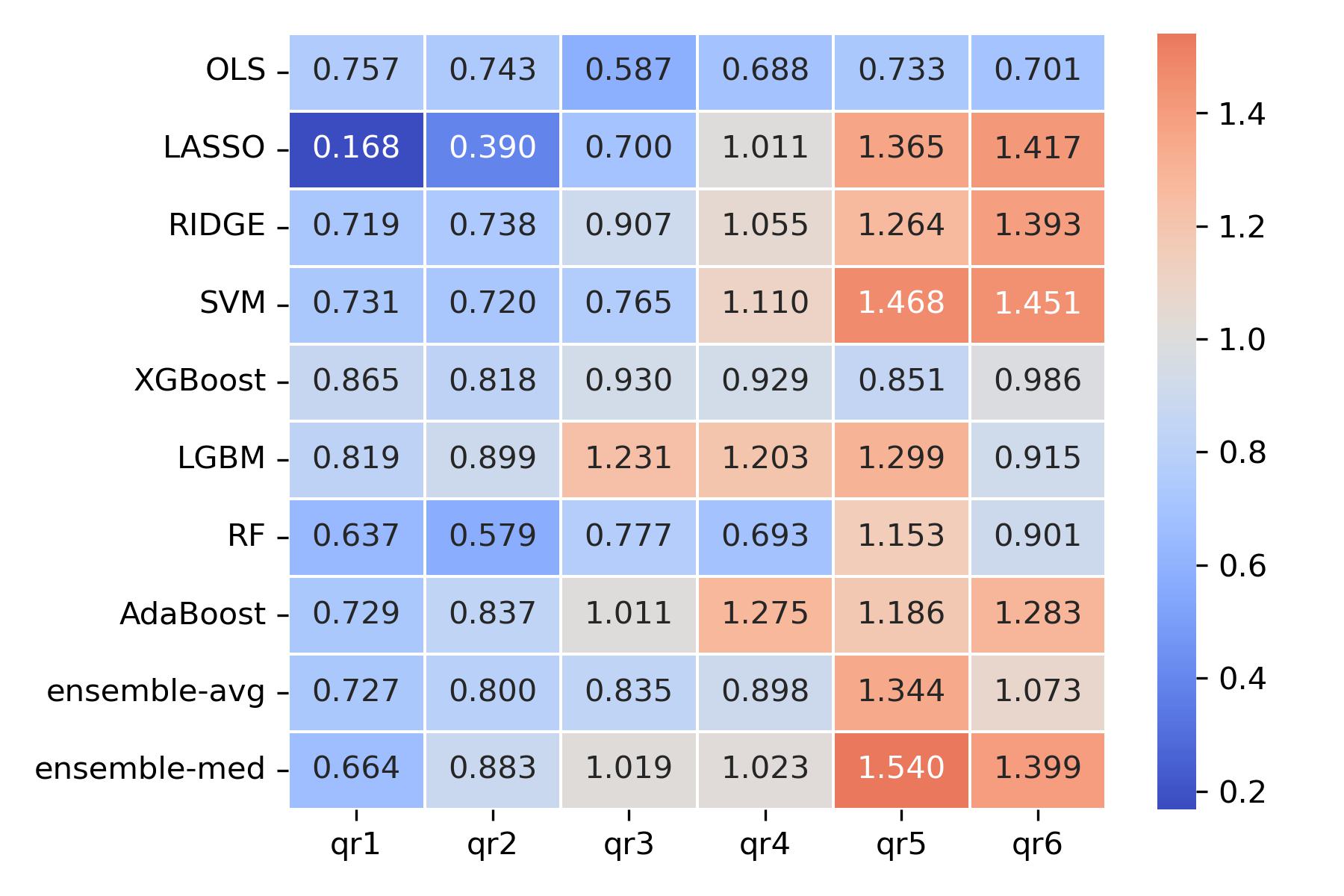}
        \caption{\mcc{Predictors: Chinese pvCLCL returns.}}
        \label{fig:CH-US_pvCLCL-OPCL_pred}
    \end{subfigure}
    \hfill 
    \begin{subfigure}{0.48\textwidth}
        \centering
        \includegraphics[width=\linewidth]{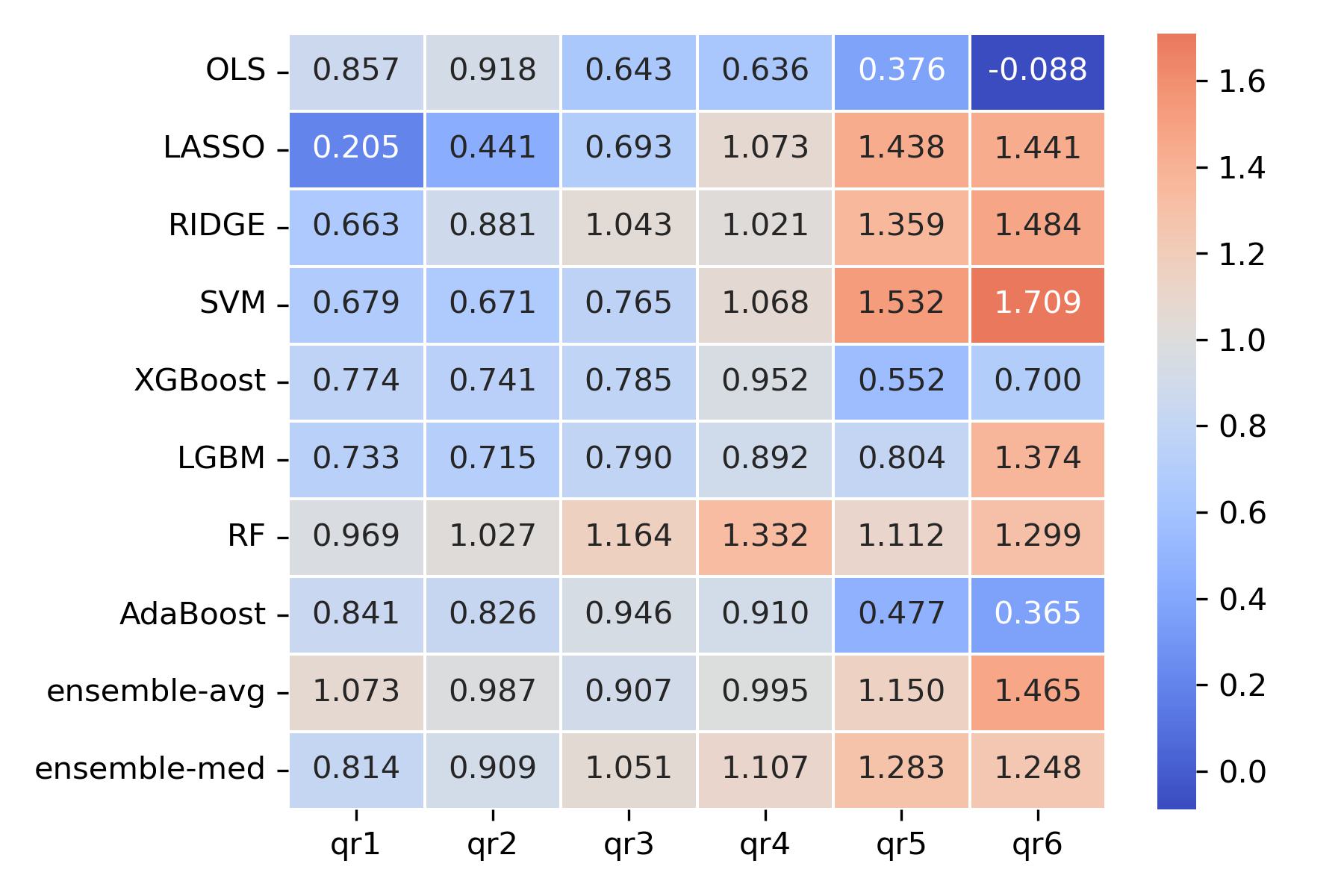}
        \caption{\mcc{Predictors: Chinese OPCL returns.}}
        \label{fig:CH-US_OPCL-OPCL_pred}
    \end{subfigure}
    \caption{\mcc{Sharpe Ratios for forecasting U.S. OPCL returns using Chinese returns as predictors.}}
    \label{fig:CH-US_pred}
\end{figure}

\begin{figure}[h!btp]
  \centering
  \includegraphics[width=0.93\linewidth]{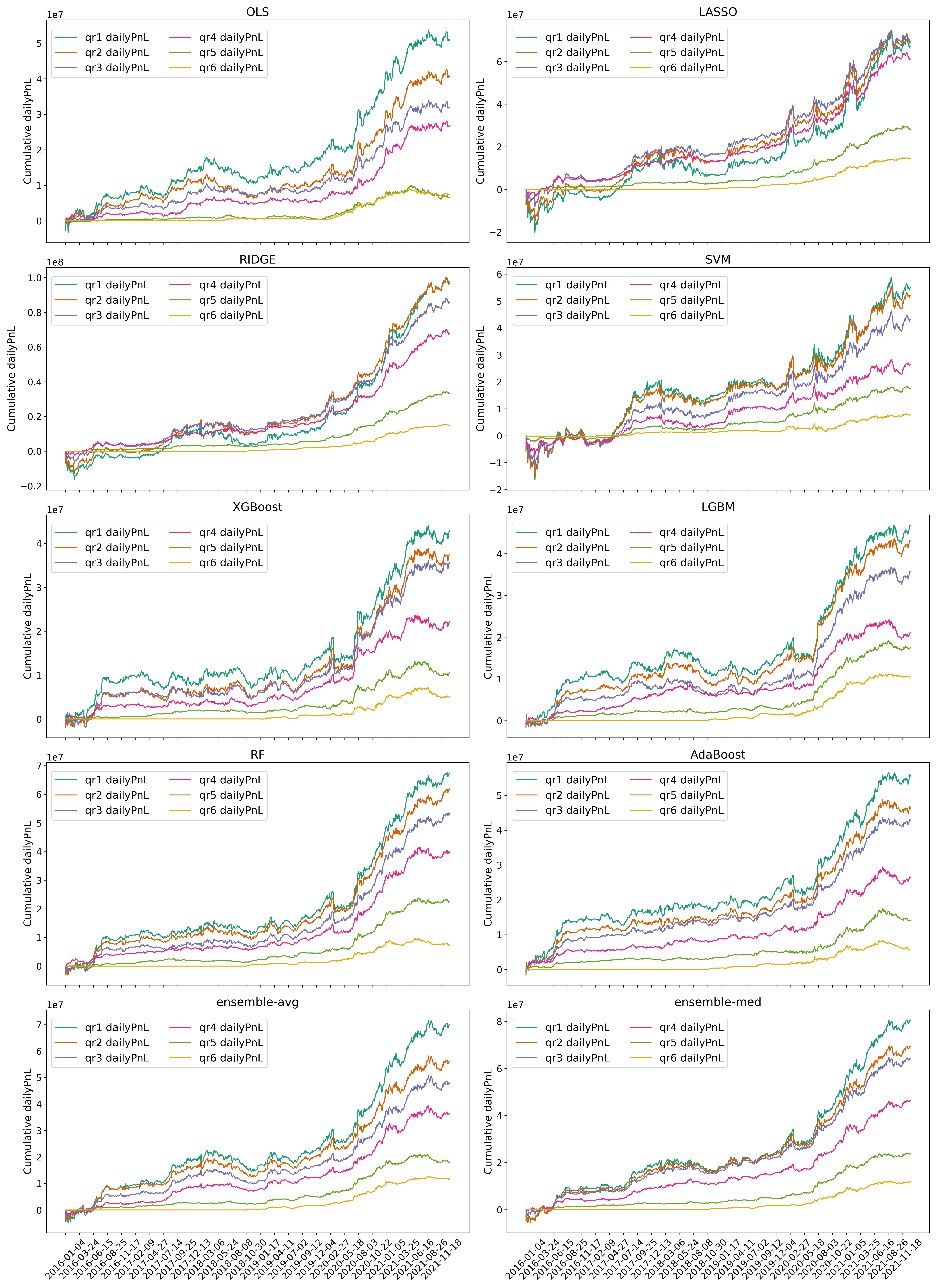}
  \vspace{-2mm}
  \caption{\mcc{Cumulative daily Profit and Loss (PnL) from forecasting Chinese OPCL returns using U.S. pvCLCL returns as predictors. Each panel corresponds to a different machine learning model, and coloured curves represent nested quantile portfolios ranked by the absolute value of predicted returns.}}  
  \label{fig:US-CH_pvCLCL-OPCL_cumPnL}
\end{figure}

Figure~\ref{fig:US-CH_pred} and Figure~\ref{fig:CH-US_pred} display the results of forecasting in two different directions. According to Figure~\ref{fig:US-CH_pred}, when predicting Chinese stocks with U.S. stocks, RIDGE, LGBM, ensemble-avg and ensemble-med yield \mcc{strong} performance. SVM \mcc{appears less effective} for this task, since its SRs are mostly lower than one. For other forecasting methods and most quantiles, SRs exceed one, with some approaching two. Notably\mcc{,} the ensemble-average and ensemble-median methods maintain robust and stable performance, often comparable to the best individual models, highlighting the benefit of model diversification. Using U.S. pvCLCL returns as features \mc{performs} better than using the U.S. OPCL returns to predict Chinese OPCL returns. The cumulative PnL plots of each method for the former are shown in Figure~\ref{fig:US-CH_pvCLCL-OPCL_cumPnL}, \mcc{where the upward-sloping trajectories indicate economically meaningful profitability}.

\mcc{In contrast,} as shown in Figure~\ref{fig:CH-US_pred}, when predicting the U.S. stocks with Chinese stocks, SRs are \mcc{substantially lower across methods and quantiles}. Therefore the Chinese market \mcc{exerts weaker predictive influence than} the U.S. market in cross-market return prediction. Since the performance is stronger when predicting Chinese stocks using U.S. pvCLCL returns, we focus on this setting in subsequent experiments and analyses.

\begin{figure}[htbp]
    \centering
    \begin{subfigure}{0.48\textwidth}
        \centering
        \includegraphics[width=\linewidth]{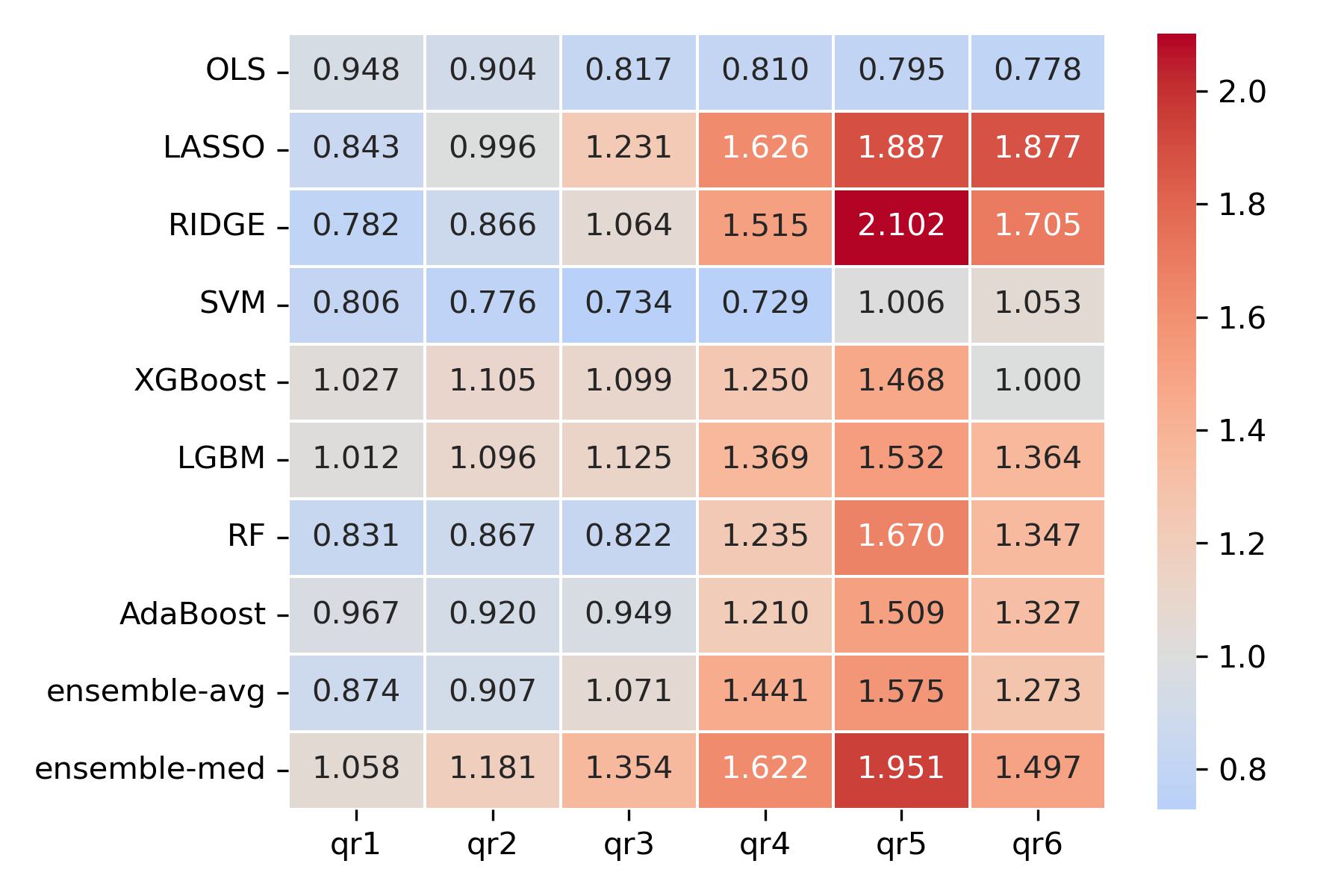}
        \caption{\mcc{Predictors: Chinese pvCLCL returns.}}
        \label{fig:CH-CH_pvCLCL-OPCL_pred}
    \end{subfigure}
    \hfill 
    \begin{subfigure}{0.48\textwidth}
        \centering
        \includegraphics[width=\linewidth]{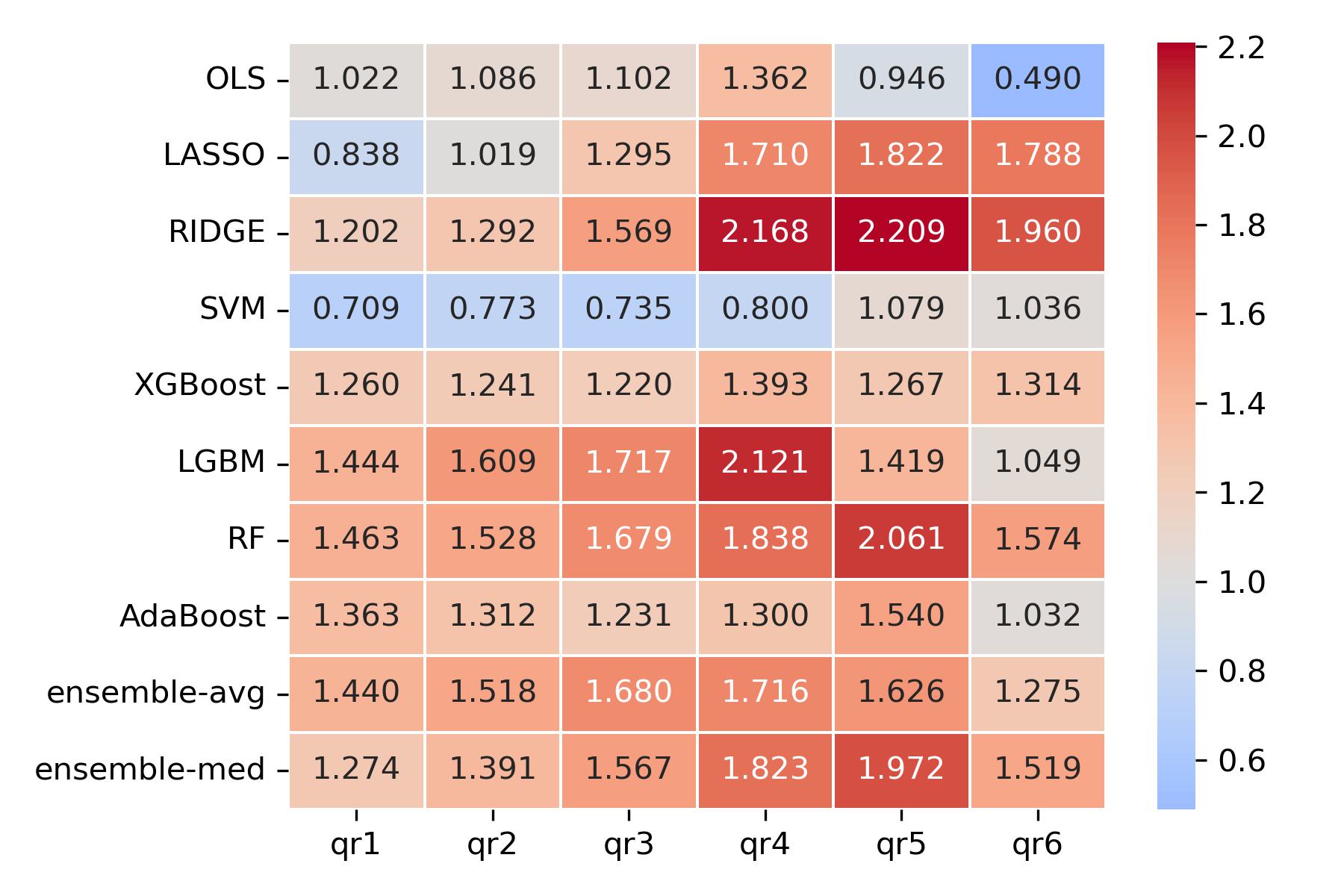}
        \caption{\mcc{Predictors: Chinese OPCL returns.}}
        \label{fig:CH-CH_OPCL-OPCL_pred}
    \end{subfigure}
  \caption{\mcc{Sharpe Ratios for forecasting Chinese OPCL returns using Chinese returns as predictors.}}    
    \label{fig:CH-CH_pred}
\end{figure}

\begin{figure}[htbp]
    \centering
    \includegraphics[width=\linewidth]{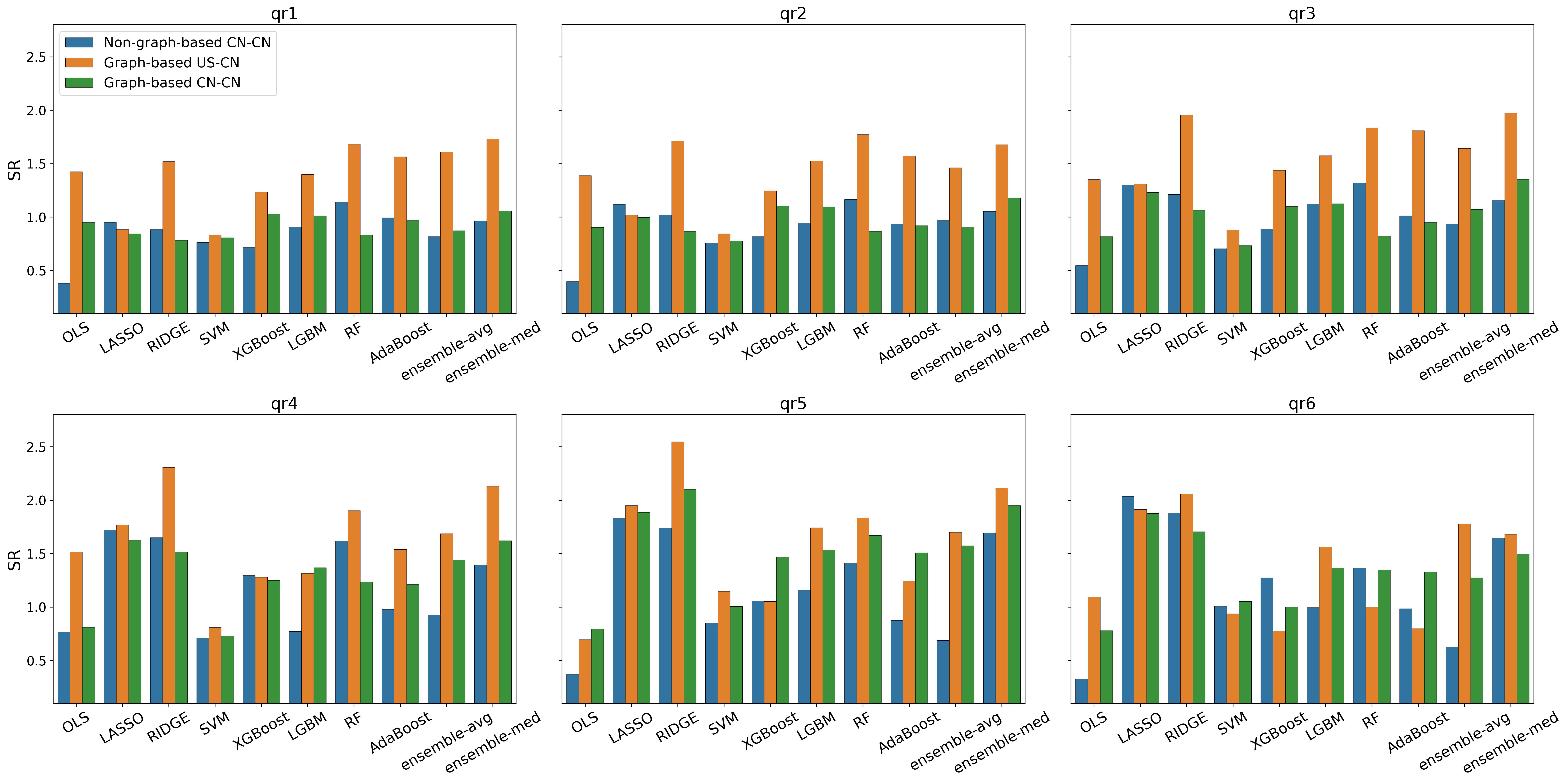}
    \caption{\mcc{Comparison of Sharpe Ratios across graph-based cross-market, graph-based same-market, and non-graph-based baseline specifications, using pvCLCL returns as predictors. CN-CN denotes using Chinese stocks to forecast Chinese stocks, while US-CN denotes using U.S. stocks to forecast Chinese stocks.}}
    \label{fig:bl_compare_3SR_pvCLCL-OPCL}
\end{figure}

\begin{figure}[htbp]
    \centering
    \includegraphics[width=0.9\linewidth]{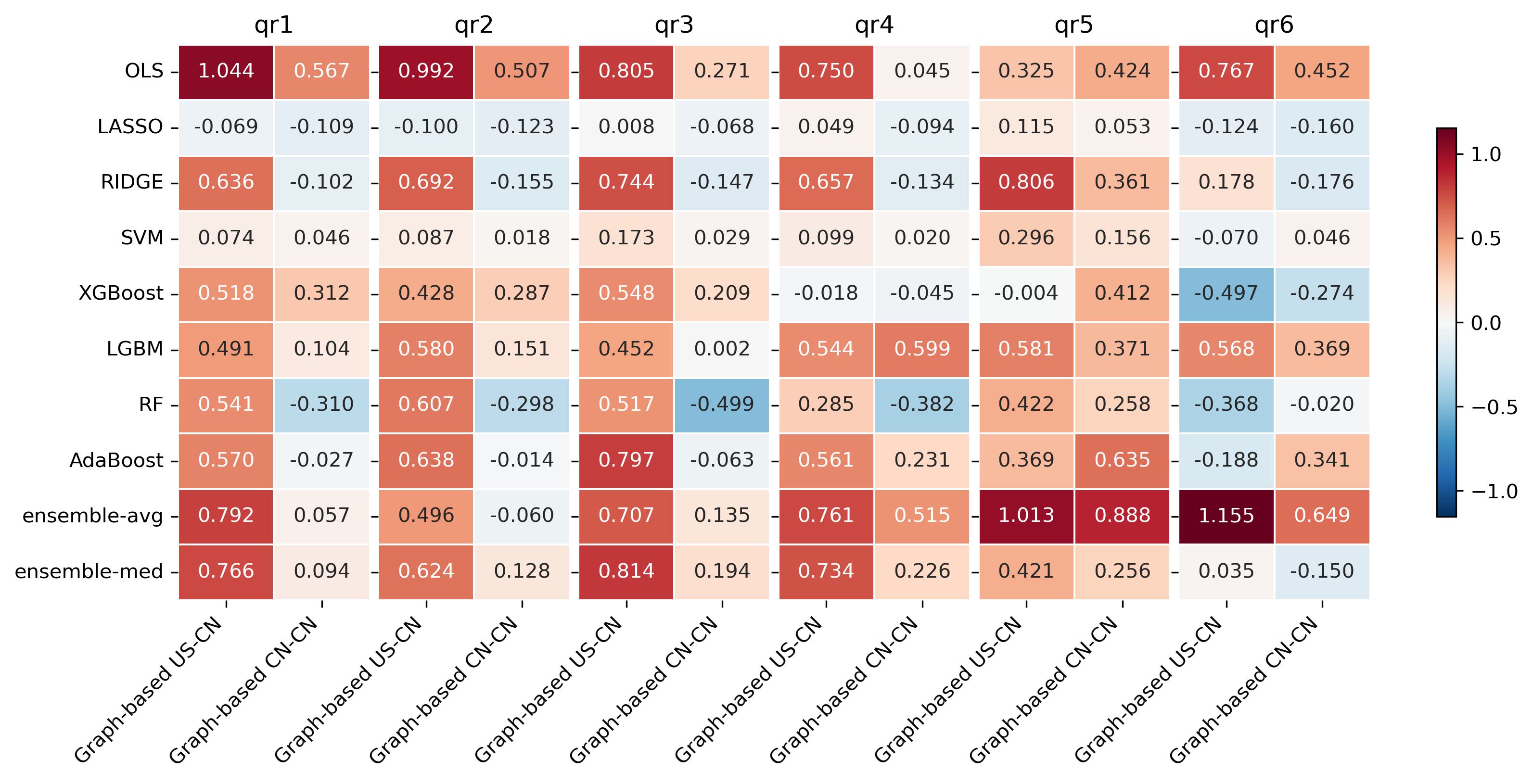}
    \caption{\mcc{Performance differentials (Sharpe Ratio deltas) relative to the non-graph-based baseline, using pvCLCL returns as predictors. US-CN denotes using U.S. stocks to forecast Chinese stocks, while CN-CN denotes using Chinese stocks to forecast Chinese stocks.}}    
    \label{fig:bl_compare_2SR_pvCLCL-OPCL}
\end{figure}

Figure~\ref{fig:CH-CH_pred} shows the results of predicting Chinese stocks with Chinese stocks based on graph structures. Figure~\ref{fig:bl_compare_3SR_pvCLCL-OPCL} summarizes SRs across specifications, comparing graph-based cross-market approaches, graph-based single-market approaches, and the non-graph-based baseline, with pvCLCL returns used as predictors. Figure~\ref{fig:bl_compare_2SR_pvCLCL-OPCL} reports the corresponding performance differentials (deltas), computed as SRs of graph-based approaches minus those of the non-graph-based baseline. The results indicate that graph-based same-market approaches outperform non-graph-based same-market approaches for most machine learning models under most quantiles, especially for OLS, LGBM, and qr5. 

\mcc{Turning to the incremental value of cross-market information,} combining cross-market information with graph information yields the strongest overall performance, outperforming approaches that use graph structures with same-market information only, as well as the non-graph-based baseline relying solely on same-market information.

\subsection{Sensitivity Analysis}
\mcc{We next evaluate the robustness of predictive performance to perturbations in graph structure and temporal alignment when forecasting Chinese returns using U.S. pvCLCL returns.}
We first conduct a feature-replacement test, where selected informative stocks are randomly substituted with other stocks. 
Additionally, we assess temporal sensitivity by varying the recency of input data, using features from earlier days (e.g., $t-2$, $t-3$, etc.) instead of the most recent day $t-1$. 
%

First, we conduct a feature-replacement experiment. Based on graphs built for predicting returns on day $t$ in the Chinese market using returns on day $t-1$ in the U.S. market, we maintain the same in-degree of each target node to preserve graph sparsity while randomly changing some of their connections. Only previously unconnected nodes are considered as replacements for the original connections. We randomly replace 20\%, 40\%, 60\%, 80\% and all of the edges. For each quantile level, we obtain the median of the results from all the 10 methods. 

\begin{figure}[htbp]
    \centering
    \begin{subfigure}[t]{0.48\textwidth}
        \centering
        \includegraphics[width=\linewidth]{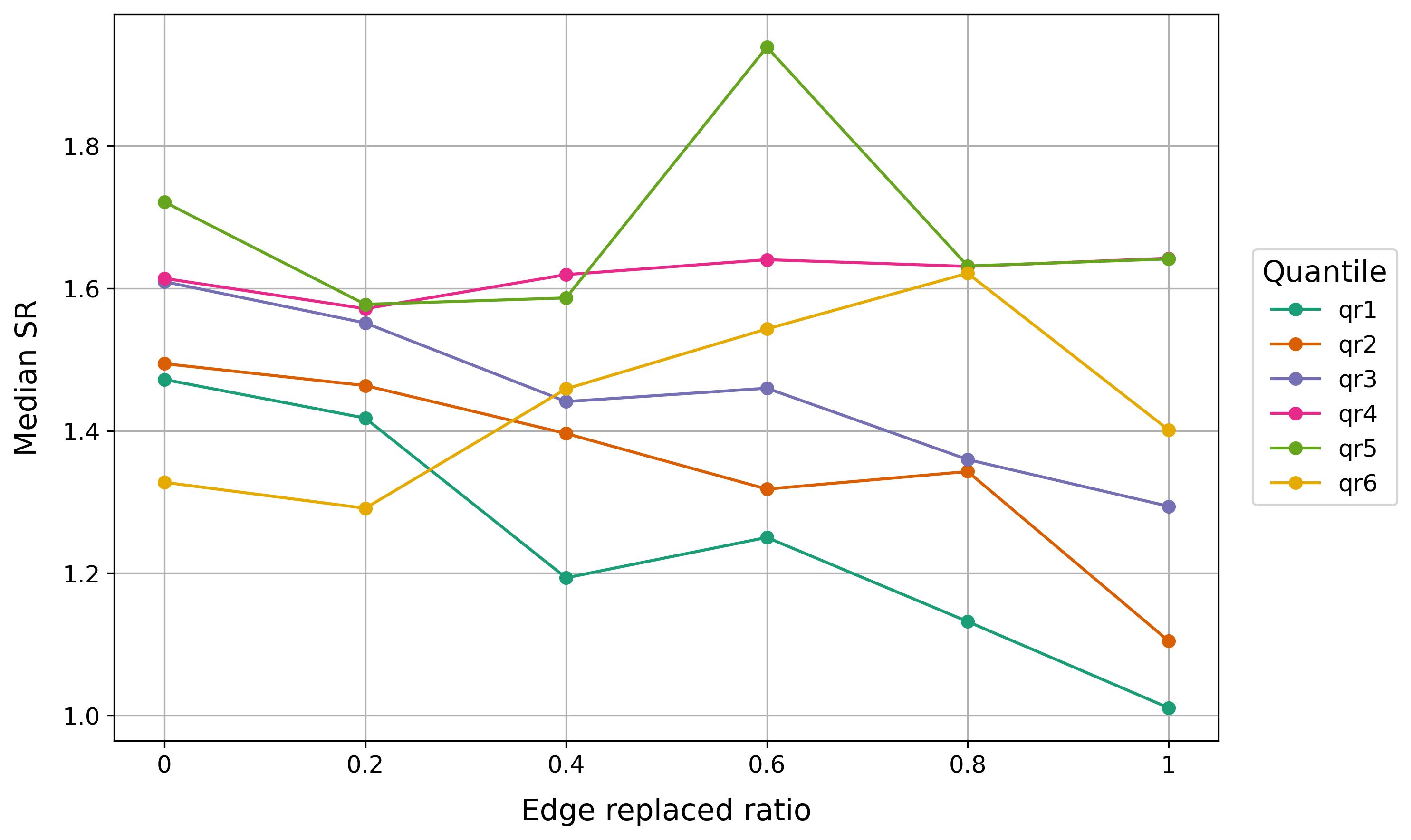}
        \caption{\mcc{Effect of graph randomization (fraction of edges replaced).}}
        \label{fig:US-CH_pvCLCL-OPCL_abrq2SR}
    \end{subfigure}    
    \hfill 
    \begin{subfigure}[t]{0.48\textwidth}
        \centering
        \includegraphics[width=\linewidth]{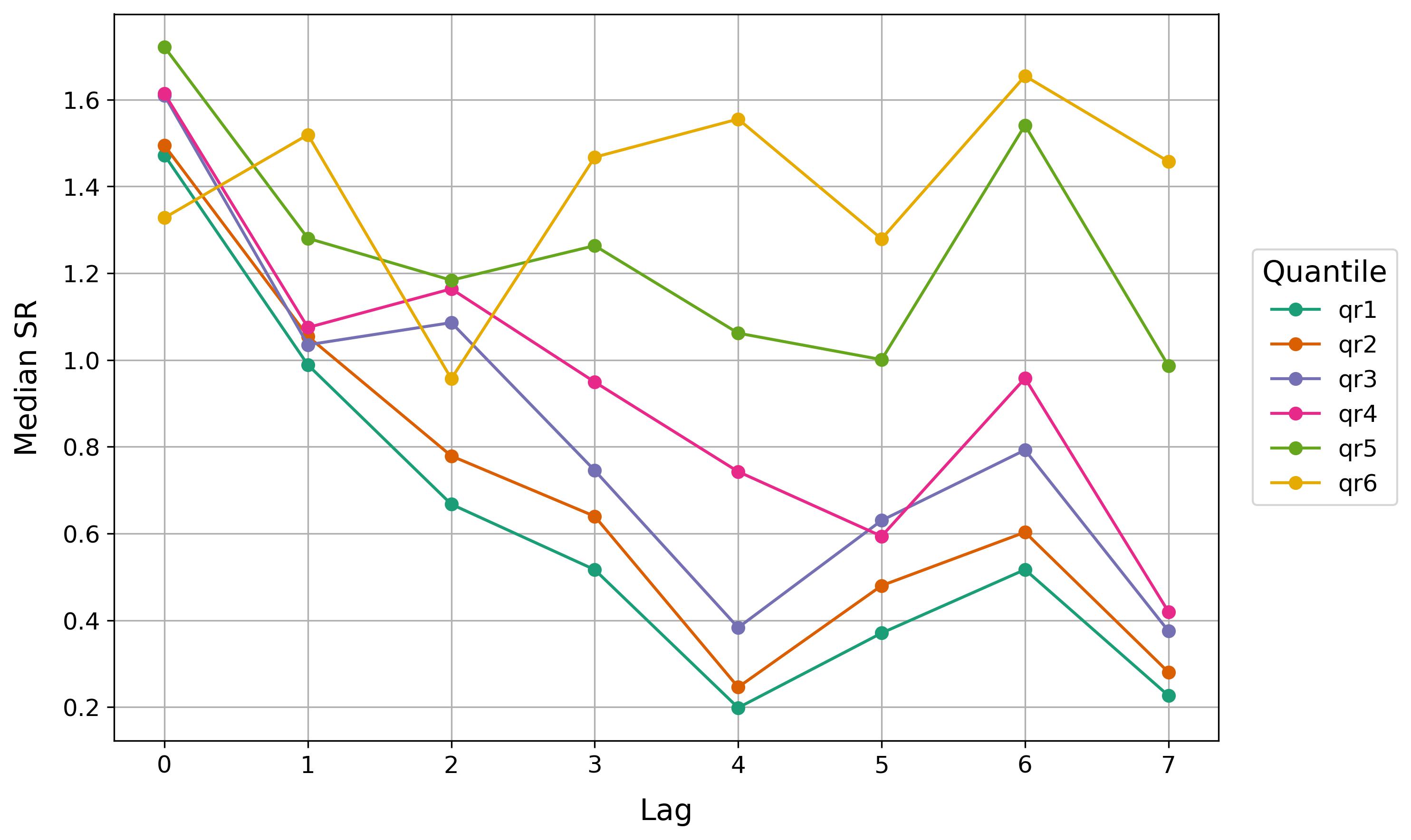}
        \caption{\mcc{Effect of increasing temporal lag $l$.}}
        \label{fig:US-CH_pvCLCL-OPCL_abRQ1oldSR}
    \end{subfigure}
    \vspace{-2mm}
    \caption{\mcc{Median forecasting performance under graph randomization (a) and increasing temporal lag (b). 
    Panel (a) reports Sharpe Ratios as a function of the fraction of replaced edges while preserving in-degree. 
    Panel (b) reports Sharpe Ratios as the lag parameter $l$ increases, measuring the effect of input recency.}}
    \label{fig:US-CH_pvCLCL-OPCL_abrq}
\end{figure}

\mcc{As shown in Figure \ref{fig:US-CH_pvCLCL-OPCL_abrq2SR}, SRs generally decline as a larger fraction of edges is replaced, indicating that predictive gains depend critically on the economically meaningful structure captured by the graph rather than on generic diversification effects. The deterioration is strongest in lower and intermediate quantiles, whereas the highest quantile (qr6) exhibits comparatively greater resilience.}

Second, we assess temporal sensitivity by varying the recency of input data. Still based on graphs built for predicting returns on day $t$ in the Chinese market using returns on day $t-1$ in the U.S. market, we look into forecasting performance as the temporal gap increases (e.g., two-day, three-day, or longer gaps) between the predictor window and the target return window. As defined in Section~\ref{sec:def}, when we forecast $r_{Y_i}^{(t)}$, \mc{the predictors are given by} $[r_{X_1}^{(t-l)}, r_{X_2}^{(t-l)}, ..., r_{X_n}^{(t-l)}]$. Here we set $l=2,3,...$ when predicting Chinese stocks. For each quantile level, we also obtain the median of the results from all the 10 methods. 

\mcc{Figure \ref{fig:US-CH_pvCLCL-OPCL_abRQ1oldSR} shows that SRs generally decline as $l$ increases, consistent with the hypothesis that cross-market predictive content decays with time. The decline is again less pronounced for qr6, suggesting that large-magnitude signals may capture more persistent cross-market effects. A mild stabilization beyond Lag 4 likely reflects weekly trading-cycle effects.}

\mcc{Taken together, these experiments confirm that predictive performance depends critically on both the structural accuracy of the graph and the recency of cross-market information.}

\section{Conclusion and Future Research}
\label{sec:end}
\mcc{This paper investigates cross-market return forecasting at the individual stock level. We develop a graph-based architecture that enables structured information transmission across markets and use it to construct cross-market predictive features. Building on this framework, we implement a range of machine learning models to forecast OPCL returns for each stock.}

\mcc{Empirically, we find that combining cross-market information with graph-based feature selection delivers superior performance relative to both graph-based same-market approaches and non-graph-based baselines. The predictive relationship is asymmetric: U.S. stocks are substantially more informative for forecasting Chinese returns than the reverse. In particular, U.S. pvCLCL returns exhibit stronger predictive power for Chinese OPCL returns than U.S. OPCL returns, highlighting the importance of overnight information transmission. Sensitivity analyses confirm that preserving the economically meaningful bipartite graph structure is crucial for achieving strong risk-adjusted performance. Moreover, forecasting performance deteriorates as the temporal gap between predictor and target returns widens, emphasizing the value of recency. }

\mcc{Several directions for future research emerge. First, extending the analysis to additional regions, including European and other Asian markets, would help assess the generalizability of cross-market predictive linkages. Second, GNNs could be applied directly to the constructed bipartite graph to learn nonlinear cross-market dependencies. Finally, recent advances in time-series-specialized large language models may offer an alternative framework for modeling structured cross-market interactions.}

\section*{Conflicts of Interest}
The authors declare that they have no competing interests.

\bibliographystyle{apalike}
\bibliography{my_ref}






\begin{appendices}
\counterwithin{figure}{section}
\counterwithin{table}{section}

\renewcommand{\thefigure}{\thesection.\arabic{figure}}
\renewcommand{\thetable}{\thesection.\arabic{table}}

\setcounter{figure}{0}
\setcounter{table}{0}


\section{Hyperparameter Settings}
The OLS method does not involve any hyperparameters. 
\mcc{Table~\ref{tab:parameter_tuning} reports the performance of the remaining seven machine learning models under alternative hyperparameter configurations.}
\mcc{The SVM model appears relatively insensitive to variations in hyperparameter values, and its SRs remain generally below one across specifications, suggesting limited suitability for this particular forecasting setting.}
\mcc{For the other models, SRs exceed one across most quantiles and hyperparameter choices, indicating that the cross-market forecasting framework delivers robust risk-adjusted performance and is not overly sensitive to specific tuning parameters.} 

\begin{sidewaystable}
\caption{Summary of SR results across different hyperparameter settings (using the most recent U.S. pvCLCL returns to predict Chinese OPCL returns). Values are reported as mean $\pm$ standard deviation.}
\label{tab:parameter_tuning}

\setlength{\tabcolsep}{10pt} 
\renewcommand{\arraystretch}{1.05}

\resizebox{\textheight}{!}{%
\begin{tabular}{ l l @{\hspace{10pt}} c c c c c c }
\toprule
\textbf{\normalsize Methods} & \textbf{\normalsize Parameters} & \textbf{\normalsize qr1} & \textbf{\normalsize qr2} & \textbf{\normalsize qr3} & \textbf{\normalsize qr4} & \textbf{\normalsize qr5} & \textbf{\normalsize qr6} \\ 
\midrule
LASSO & $\lambda$=\{1e-4, 1e-3, 1e-2, 1e-1, 1, 10, 100, 1000\} & $0.848\pm0.222$ & $0.969\pm0.247$ & $1.238\pm0.31$ & $1.65\pm0.398$ & $1.821\pm0.433$ & $1.777\pm0.424$ \\
\midrule
RIDGE & $\lambda$=\{1e-4, 1e-3, 1e-2, 1e-1, 1, 10, 100, 1000\} & $1.238\pm0.429$ & $1.293\pm0.43$ & $1.437\pm0.449$ & $1.715\pm0.51$ & $1.668\pm0.619$ & $1.55\pm0.461$ \\
\midrule
SVM & $C$=\{0.1, 1, 10, 100, 1000\} & $0.743\pm0.194$ & $0.751\pm0.196$ & $0.782\pm0.204$ & $0.719\pm0.188$ & $1.02\pm0.267$ & $0.836\pm0.218$ \\
\midrule
XGBoost & \begin{tabular}[c]{@{}l@{}}$\text{max\_depth}$ = \{3,6,9\}\\ $\text{learning\_rate}$ = \{0.01, 0.1, 0.2\}\\ $\text{n\_estimators}$ = \{50, 100, 300\}\end{tabular} & $1.445\pm0.224$ & $1.48\pm0.247$ & $1.587\pm0.214$ & $1.539\pm0.309$ & $1.381\pm0.401$ & $0.964\pm0.233$ \\
\midrule
LGBM & \begin{tabular}[c]{@{}l@{}}$\text{num\_leaves}$ = \{10, 31, 90\}\\ $\text{learning\_rate}$ = \{0.01, 0.1, 0.2\}\\ $\text{n\_estimators}$ = \{50, 100, 300\}\end{tabular} & $1.32\pm0.348$ & $1.457\pm0.388$ & $1.541\pm0.422$ & $1.44\pm0.485$ & $1.652\pm0.466$ & $1.424\pm0.407$ \\
\midrule
RF & \begin{tabular}[c]{@{}l@{}}$\text{n\_estimators}$ = \{50, 100, 300\}\\ $\text{max\_depth}$ = \{5, 10, 50\}\end{tabular} & $1.481\pm0.41$ & $1.558\pm0.422$ & $1.639\pm0.447$ & $1.799\pm0.486$ & $1.747\pm0.474$ & $1.09\pm0.351$ \\
\midrule
AdaBoost & \begin{tabular}[c]{@{}l@{}}$\text{max\_depth}$ = \{1, 5, 10\}\\ $\text{n\_estimators}$ = \{50, 100, 300\}\\ $\text{learning\_rate}$ = \{0.01, 0.1, 0.2\}\end{tabular} & $1.389\pm0.358$ & $1.405\pm0.385$ & $1.549\pm0.436$ & $1.548\pm0.419$ & $1.1\pm0.305$ & $0.667\pm0.187$ \\ \bottomrule
\end{tabular}}
\begin{tablenotes}%
\item Note: $\text{max\_depth}$ is the maximum depth of a tree.
$\text{learning\_rate}$ is the learning rate.
$\text{n\_estimators}$ is the maximum number of estimators at which boosting is terminated.
$\text{num\_leaves}$ is maximum tree leaves for base learners in an LGBM model.

\end{tablenotes}
\end{sidewaystable}

\section{Comparison with Baselines when Predicting with OPCL Returns}
\begin{figure}[htbp]
    \centering
    \includegraphics[width=\linewidth]{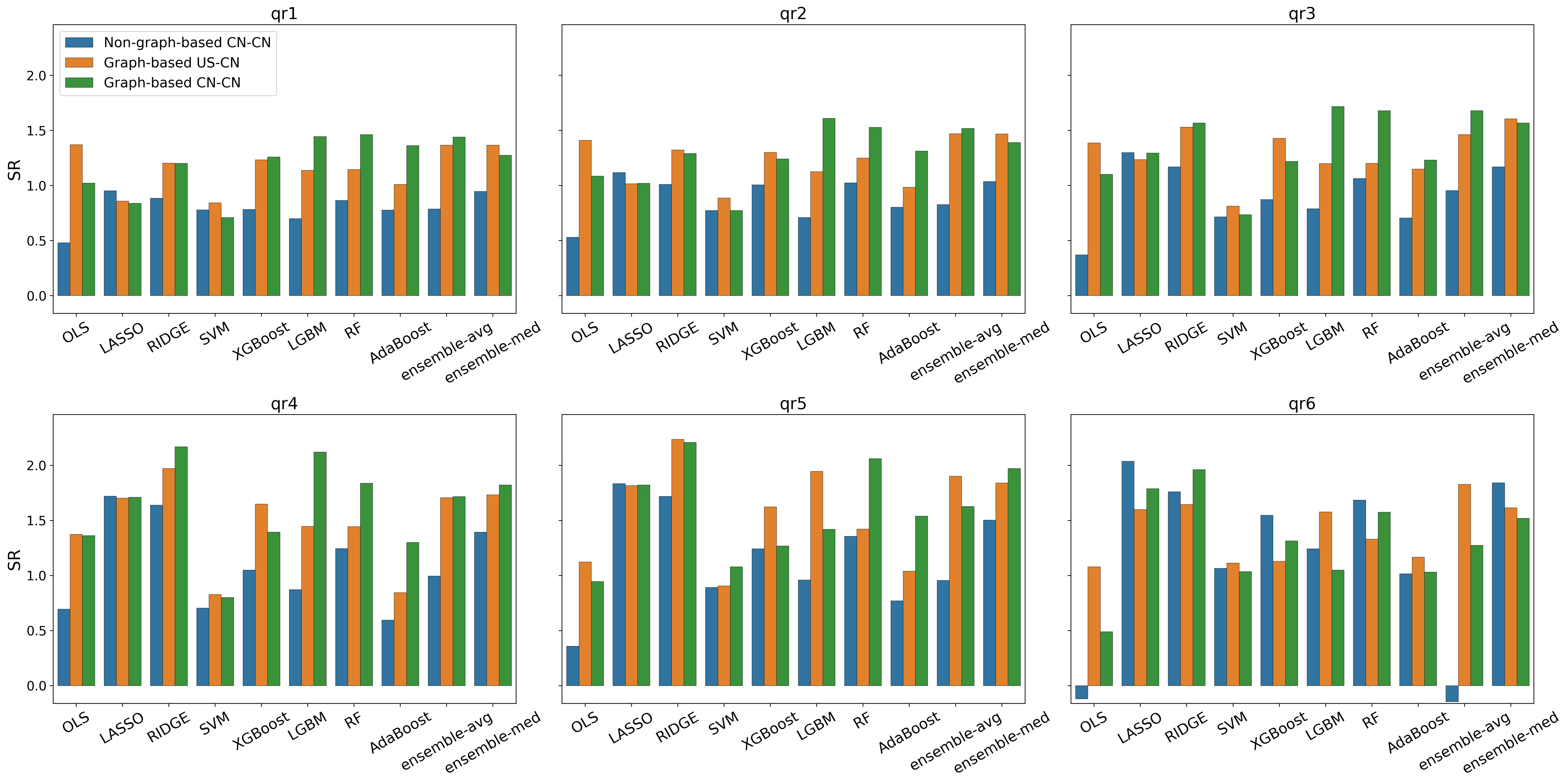}
    \caption{Comparison of Sharpe Ratios across graph-based cross-market, graph-based same-market, and non-graph-based baseline specifications, using OPCL returns as predictors. US-CN denotes using U.S. stocks to forecast Chinese stocks, while CN-CN denotes using Chinese stocks to forecast Chinese stocks.}
    \label{fig:bl_compare_3SR_OPCL-OPCL}
\end{figure}

\begin{figure}[htbp]
    \centering
    \includegraphics[width=0.95\linewidth]{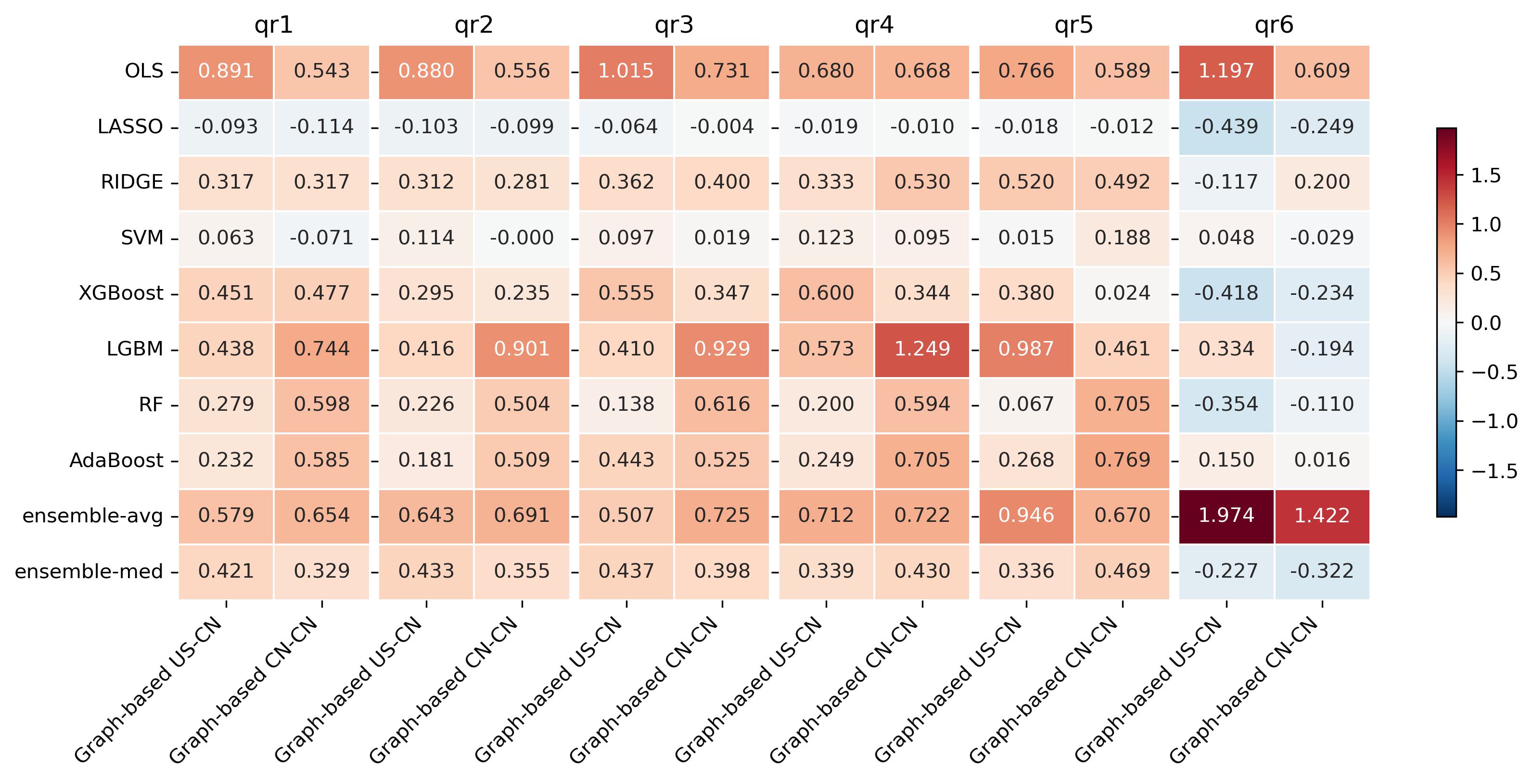}
\caption{Performance differentials (Sharpe Ratio deltas) relative to the non-graph-based baseline, using OPCL returns as predictors. US-CN denotes using U.S. stocks to forecast Chinese stocks, while CN-CN denotes using Chinese stocks to forecast Chinese stocks.}
    \label{fig:bl_compare_2SR_OPCL-OPCL}
\end{figure}

Similar to Figure~\ref{fig:bl_compare_3SR_pvCLCL-OPCL} and Figure~\ref{fig:bl_compare_2SR_pvCLCL-OPCL}, Figure~\ref{fig:bl_compare_3SR_OPCL-OPCL} and Figure~\ref{fig:bl_compare_2SR_OPCL-OPCL} \mcc{report the performance comparison among the graph-based cross-market approach, the graph-based same-market baseline, and the non-graph-based baseline when OPCL returns are used as predictors.}

\mcc{The results indicate that incorporating graph structure generally improves forecasting performance across most quantiles and machine learning models. However, in contrast to the pvCLCL setting, the incremental contribution of cross-market information is substantially weaker when OPCL returns are used as predictors.}

\section{\mc{Cross-Market Spillovers under U.S. Market Shocks}} 
\mcc{To investigate whether cross-market predictability strengthens during periods of heightened U.S. market activity and volatility, we condition the analysis on large movements in the U.S. market. Specifically, we identify days on which the S\&P 500 ETF (SPY) exhibits large absolute pvCLCL returns.}

We divide the SPY absolute pvCLCL returns into \mcc{nested} quantiles (qr1–qr6), consistent with the earlier quantile construction. For example, qr2 contains the top 80\% of days ranked by absolute SPY returns. For each identified SPY shock date $t_i$, we extract the model’s OPCL return forecast for the next trading day $t_i + 1$ in the Chinese market. We then compute SRs restricted to these subsets of dates. \mcc{This conditional evaluation allows us to assess whether predictive performance improves during periods of elevated U.S. market volatility, when cross-market information transmission is likely to be stronger.}

\begin{figure}[htbp]
\centering
\includegraphics[width=0.5\linewidth]{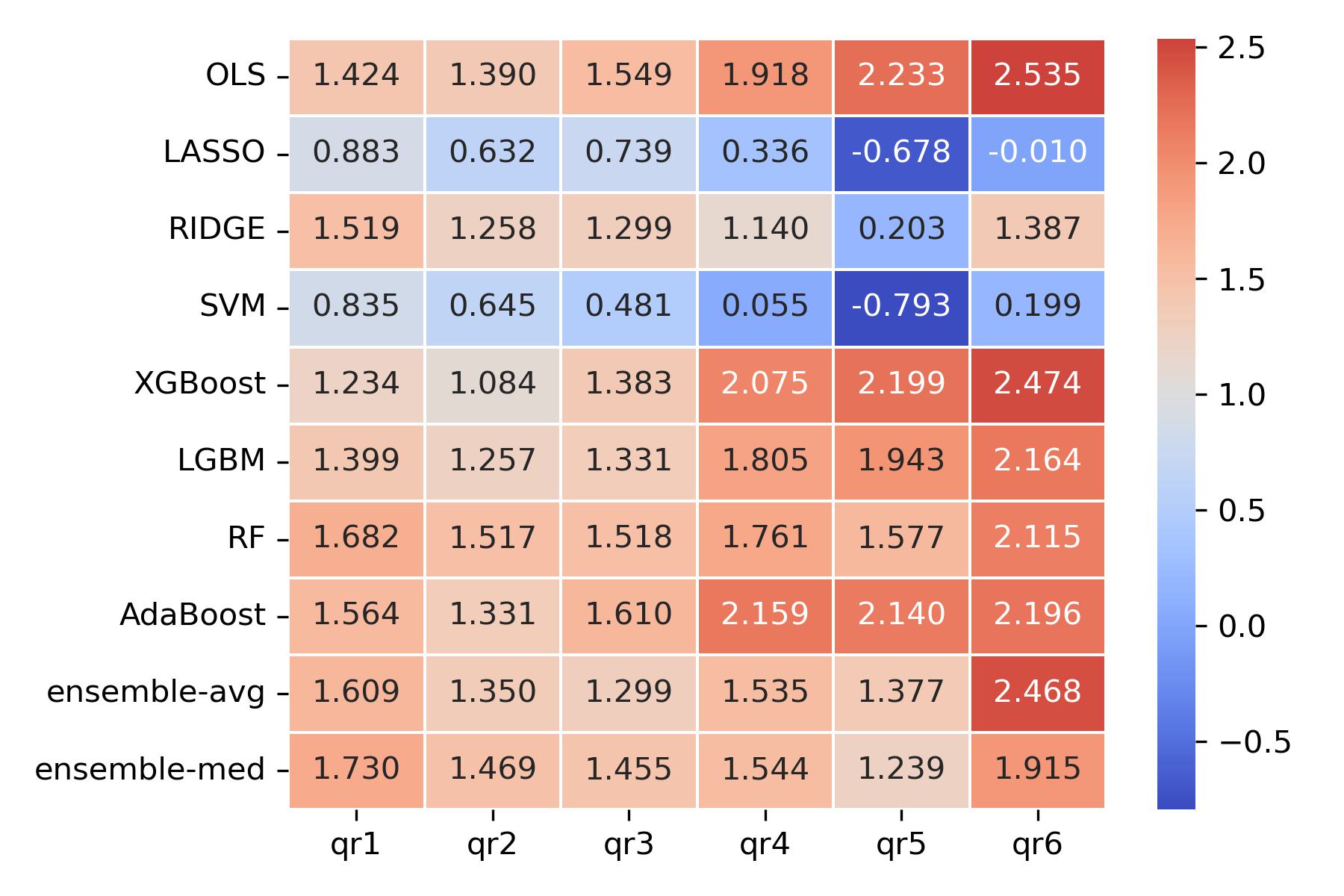}
\caption{\mcc{Sharpe Ratios of forecasting models conditional on SPY absolute pvCLCL return nested quantiles, measuring predictive performance following different magnitudes of U.S. market shocks.}}
\label{fig:spillover}
\end{figure}

The results are shown in Figure~\ref{fig:spillover}. Nonlinear ensemble models such as XGBoost, LGBM, RF, and AdaBoost achieve consistently high SRs across quantiles, especially in the upper ranges (qr4–qr6), suggesting that \mcc{nonlinear models capture state-dependent spillover effects more effectively}. OLS also performs well, achieving its highest SR in qr6. In contrast, LASSO and SVM exhibit weaker performance, with SRs turning negative in some quantiles, while RIDGE shows comparatively unstable performance.

\section{Comparison of Sharpe Ratio Results across Sectors}
\begin{figure}[t]
\centering
\includegraphics[width=\linewidth]{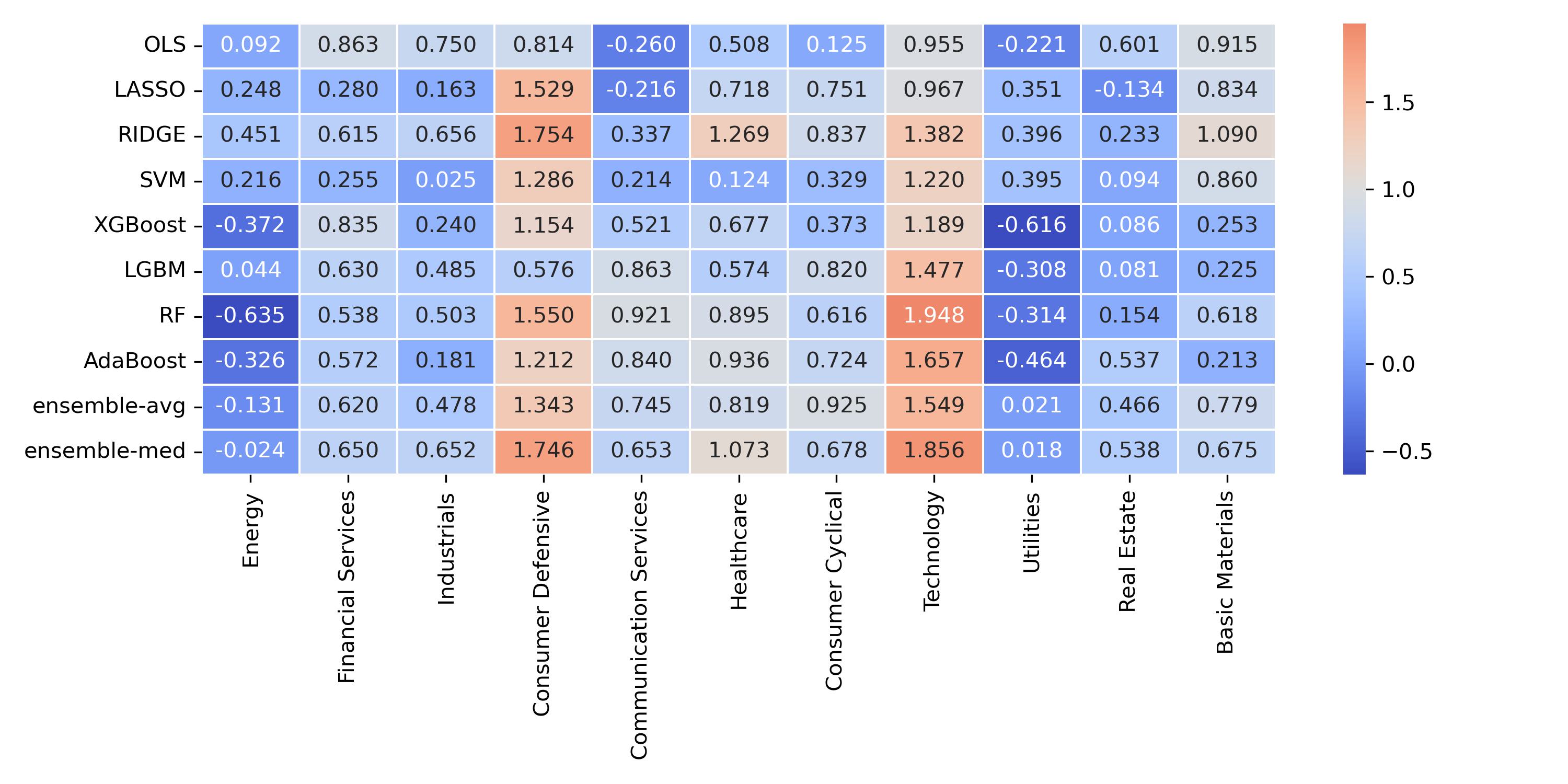}
\caption{\mcc{Sector-level Sharpe Ratios for cross-market forecasting (using U.S. pvCLCL returns to forecast Chinese OPCL returns).}}
\label{fig:sec_sompare_SR}
\end{figure}

\mcc{We continue to focus on the setting in which U.S. pvCLCL returns are used to forecast Chinese OPCL returns.} The sector-level SR results are reported in Figure \ref{fig:sec_sompare_SR}. \mcc{Because the number of stocks varies across sectors, these results should be interpreted with caution. Overall, the technology and consumer defensive sectors exhibit relatively higher SRs compared with other sectors, indicating stronger cross-market predictive effects in these segments.}

\section{Predicting Chinese Stocks Using Both U.S. and Chinese Stocks}
\begin{figure}[!t]
    \centering

    \begin{subfigure}{0.48\textwidth}
        \centering
        \includegraphics[width=\linewidth]{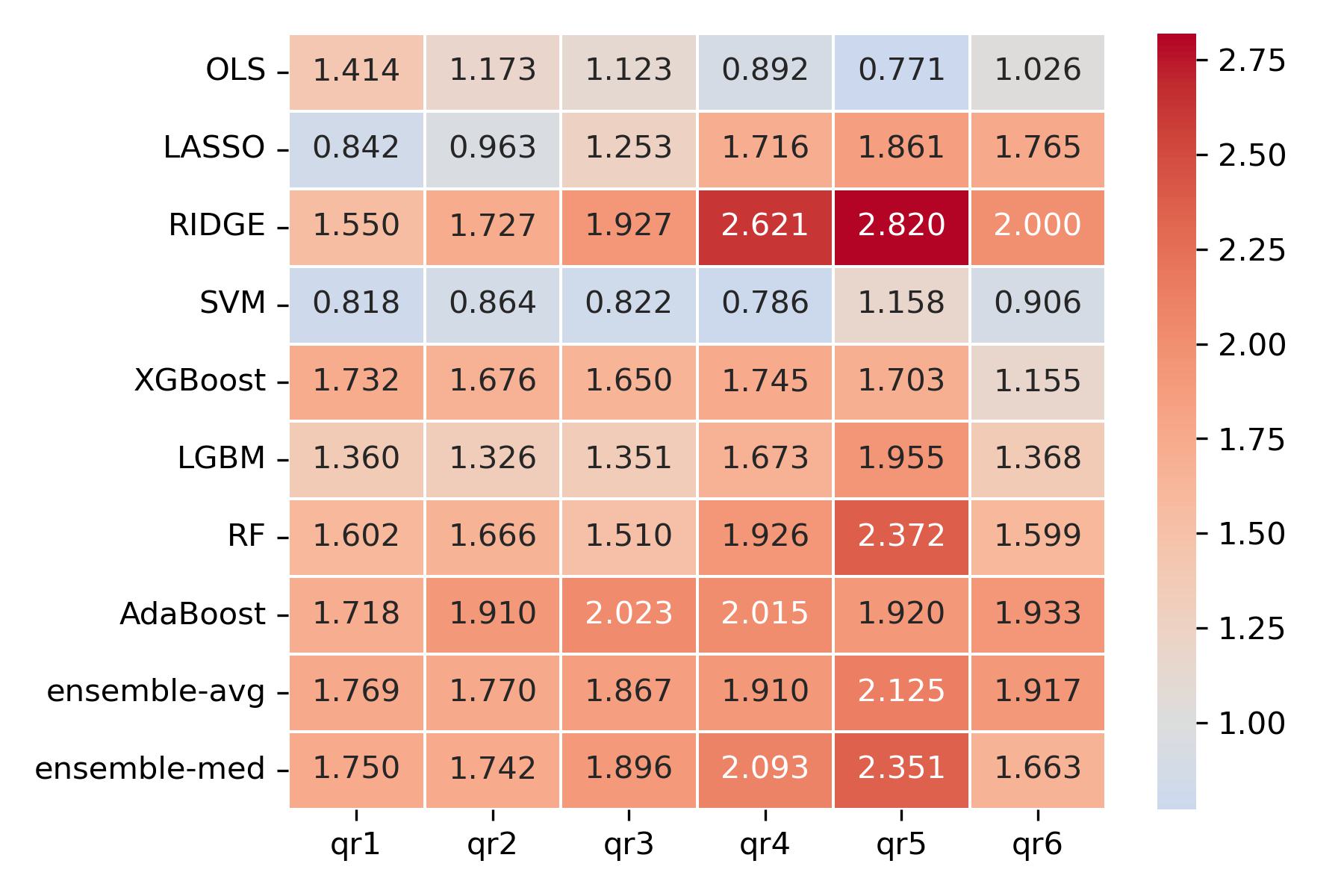}
        \caption{U.S. pvCLCL + Chinese pvCLCL returns.}
        \label{fig:US+CH-CH_pvCLCL+pvCLCL-OPCL_pred}
    \end{subfigure} 
    \hfill
    \begin{subfigure}{0.48\textwidth}
        \centering
        \includegraphics[width=\linewidth]{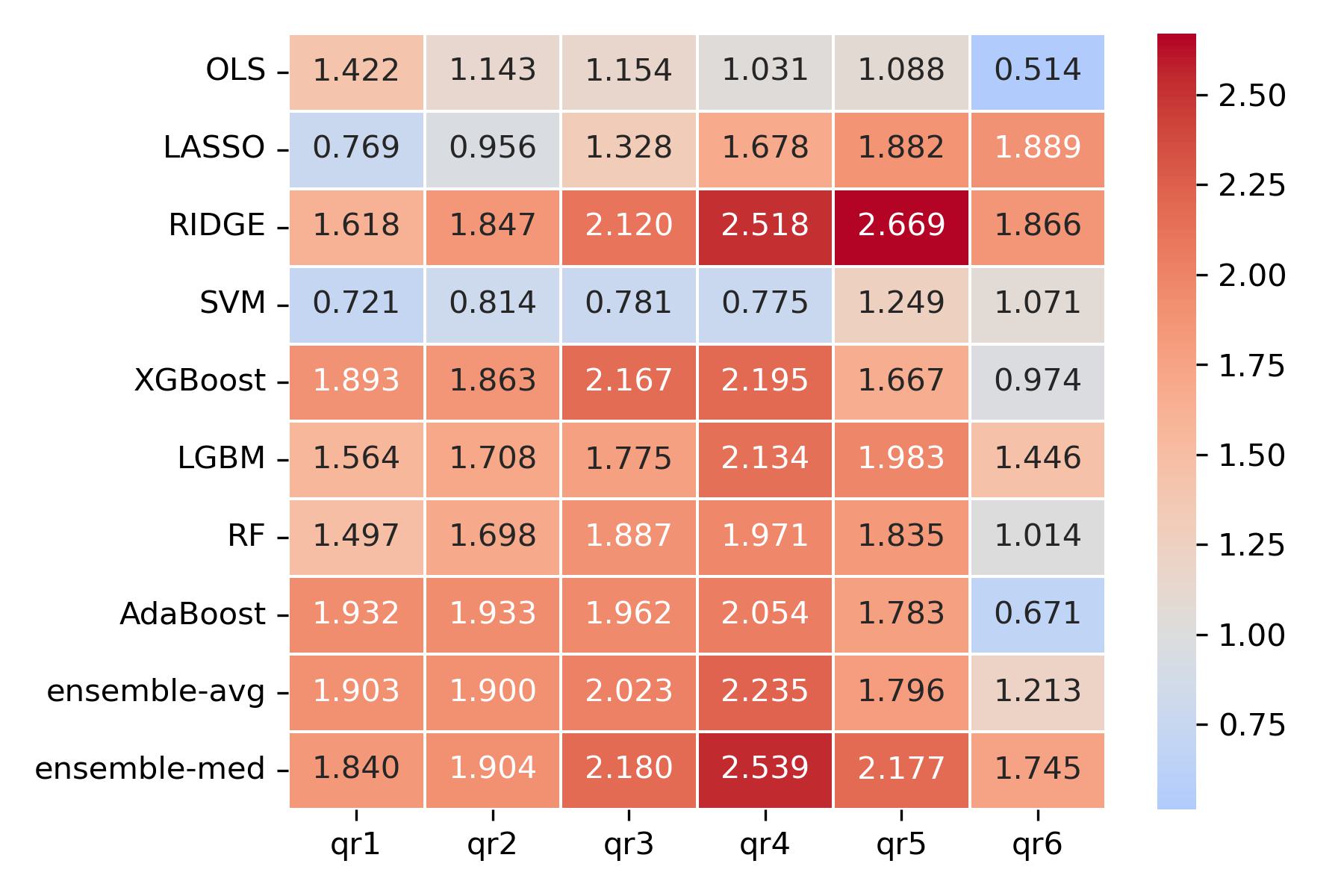}
        \caption{U.S. pvCLCL + Chinese OPCL returns.}
        \label{fig:US+CH-CH_pvCLCL+OPCL-OPCL_pred}
    \end{subfigure}
    \vspace{0.5em} 
    \begin{subfigure}{0.48\textwidth}
        \centering
        \includegraphics[width=\linewidth]{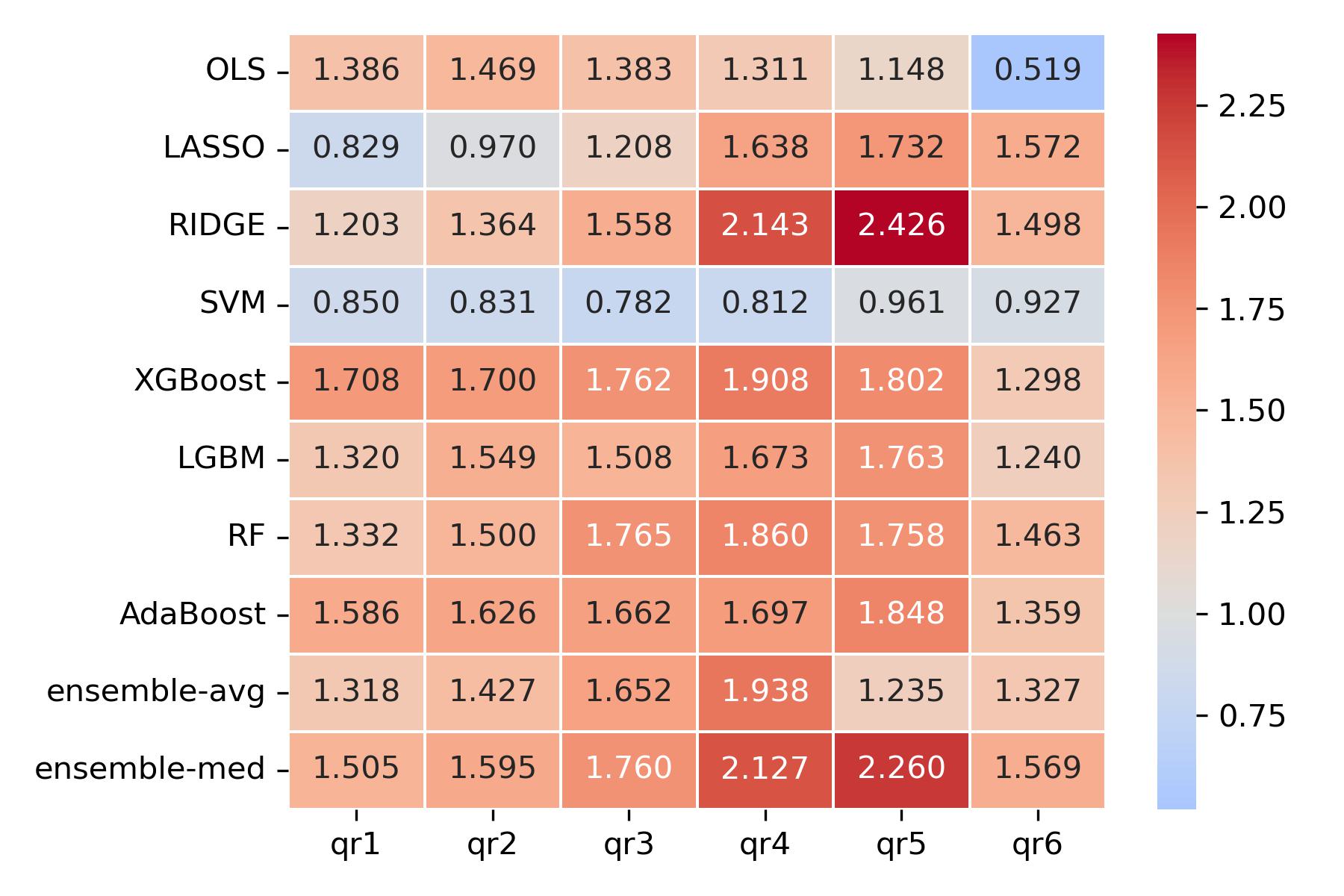}
        \caption{U.S. OPCL + Chinese pvCLCL returns.}
        \label{fig:US+CH-CH_OPCL+pvCLCL-OPCL_pred}
    \end{subfigure}
    \hfill
    \begin{subfigure}{0.48\textwidth}
        \centering
        \includegraphics[width=\linewidth]{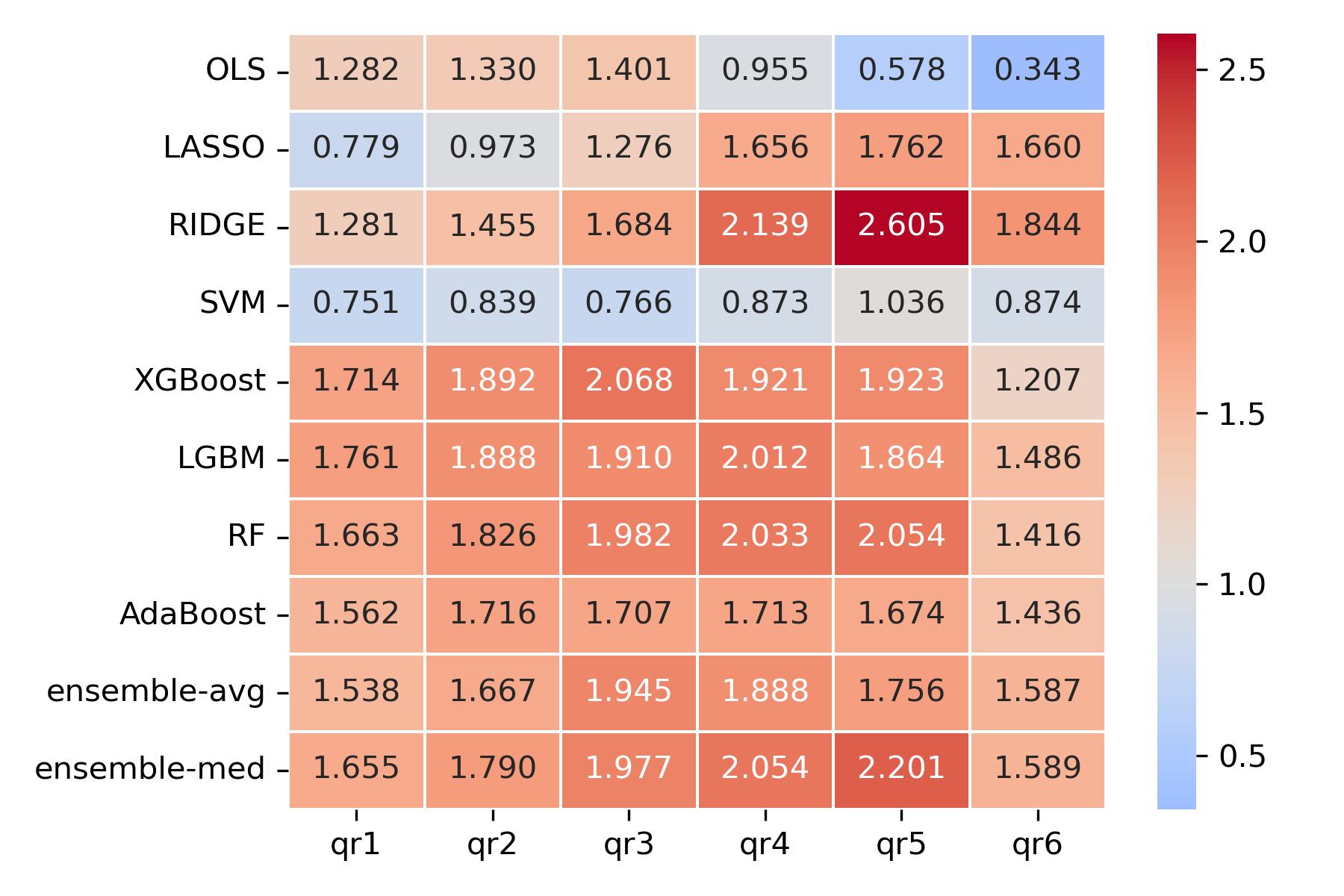}
        \caption{U.S. OPCL + Chinese OPCL returns.}
        \label{fig:US+CH-CH_OPCL+OPCL-OPCL_pred}
    \end{subfigure}
\caption{\mcc{Sharpe Ratios for forecasting Chinese OPCL returns when the predictor set includes both U.S. and Chinese stocks. Panels vary the combination of pvCLCL and OPCL features across the two markets.}}
\label{fig:US+CH-CH_pred}
\end{figure}

\mcc{We extend the analysis by allowing the predictor set $\mathcal{X}$ to include all stocks from both the U.S. and Chinese markets, while the target set $\mathcal{Y}$ continues to consist of Chinese stocks only.} The graph construction phase follows the procedure introduced in Section \ref{sec:debigraph}. The results are presented in Figure \ref{fig:US+CH-CH_pred}.  \mcc{In several specifications, the resulting SRs exceed those obtained when using only U.S. returns as predictors (Figure \ref{fig:US-CH_pred}), suggesting that combining domestic and cross-market information can further enhance predictive performance.}

\clearpage
\end{appendices}

\end{document}